\documentclass{svjour3}
\usepackage{amsmath}
\smartqed
\usepackage{amssymb}
\usepackage[justification=centering]{caption}
\usepackage[a4paper, total={7in, 10in}]{geometry}
\usepackage{graphicx}
\graphicspath{{visuals/}}
\usepackage{hyperref}
\hypersetup{
    colorlinks,
    citecolor=blue,
    filecolor=blue,
    linkcolor=blue,
    urlcolor=blue
}

\usepackage[utf8]{inputenc}
\usepackage{lipsum, natbib, pdfpages, placeins, todonotes, url, soul, booktabs}
\usepackage{colortbl}
\usepackage{comment}
\usepackage{subcaption}
\usepackage{textcomp}
\usepackage{tipa}
\newcommand{\element}[1]{\noindent\textbf{#1} ~~}
\newcommand{\ie}[1]{\textit{i.e.}~#1}
\newcommand{\eg}[1]{\textit{e.g.}~#1}

\newcommand{\tedtrans}{IWSLT~’18}
\newcommand{\vatex}{\textsc{VaTeX}}
\newcommand{\taskf}[1]{\rotatebox{00}{\textbf{\textsc{#1}}}}
\newcommand{\tick}{\checkmark}

\newcommand{\lp}[2]{#1$\rightarrow$#2}

\definecolor{unc}{HTML}{eeeeee}
\definecolor{ssm}{HTML}{bf0808}
\definecolor{ssp}{HTML}{2c7a14}
\newcommand{\uua}[0]{$\uparrow\,$}
\newcommand{\dda}[0]{$\mathbf{\downarrow}\,$}

\newcommand{\dg}[0] {$\,\dagger$}

\newcommand{\sspn}[1]{\textcolor{ssp}{(\uua #1)}}
\newcommand{\ssmn}[1]{\textcolor{ssm}{(\dda #1)}}
\newcommand{\noc}[0]{\textcolor{gray}{(-- 0.0)}}
\newcommand{\na}[0]{\textcolor{gray}{(N/A)}}

\newcommand{\ff}[0]{$\bigstar$}

\def\spatfeats{{\mathbf{V}}}    
\def\globfeats{{v}}             
\newcommand{\R}{\mathbb{R}}     
\newcommand{\argmax}{\mathop{\mathrm{argmax}}\limits} 

\title{
    Multimodal Machine Translation through Visuals and Speech
}

\author{
    Umut Sulubacak \and Ozan Caglayan \and Stig-Arne Grönroos \\
    Aku Rouhe \and Desmond Elliott \and Lucia Specia \and Jörg Tiedemann
}

\institute{
    \begin{minipage}[t]{0.5\textwidth}
        Umut Sulubacak \and Jörg Tiedemann \at University of Helsinki \\
            \email{\{umut.sulubacak\,$|$\,jorg.tiedemann\}\,@\,helsinki.fi} \and
        Stig-Arne Grönroos \and Aku Rouhe \at Aalto University \\
            \email{\{stig-arne.gronroos\,$|$\,aku.rouhe\}\,@\,aalto.fi}
    \end{minipage}
    \begin{minipage}[t]{0.5\textwidth}
        Ozan Caglayan \and Lucia Specia \at Imperial College London \\
            \email{\{o.caglayan\,$|$\,l.specia\}\,@\,imperial.ac.uk} \and
        Desmond Elliott \at University of Copenhagen \\
            \email{de\,@\,di.ku.dk}
    \end{minipage}
}

\date{}
\begin{document}
\maketitle

\abstract{
    
    Multimodal machine translation involves drawing information from more than one modality, based on the assumption that the additional modalities will contain useful alternative views of the input data. The most prominent tasks in this area are spoken language translation, image-guided translation, and video-guided translation, which exploit audio and visual modalities, respectively. These tasks are distinguished from their monolingual counterparts of speech recognition, image captioning, and video captioning by the requirement of models to generate outputs in a different language. This survey reviews the major data resources for these tasks, the evaluation campaigns concentrated around them, the state of the art in end-to-end and pipeline approaches, and also the challenges in performance evaluation. The paper concludes with a discussion of directions for future research in these areas: the need for more expansive and challenging datasets, for targeted evaluations of model performance, and for multimodality in both the input and output space.
}


\section{Introduction}
\label{sec:intro}
    Humans are able to make use of complex combinations of visual, auditory, tactile and other stimuli, and are capable of not only handling each sensory modality in isolation, but also
    simultaneously integrating them to improve the quality of perception and understanding~\citep{stein-neural-basis-2009}.
    From a computational perspective, natural language processing~(NLP) requires such abilities, too, in order to approach human-level grounding and understanding in various AI tasks.
    
    While language covers written, spoken, and sign language in human communication; vision, speech, and language processing communities have worked largely apart in the past. As a consequence, NLP became more focused towards \emph{textual} representations, which often disregard many other characteristics of communication such as
    non-verbal auditory cues, facial expressions, and hand gestures.
    Luckily, recent advances in multimodal machine learning have brought these different aspects of language together, through a plethora of multimodal NLP tasks.
    Specifically, these tasks involve more than one modality, either by (i) using one modality to aid the interpretation of language in another modality, or by (ii) converting one modality into another. Notable examples for the first category are extensions to initially unimodal problems, such as multimodal coreference resolution~\citep{ramanathan2014linking}, multimodal sentiment analysis~\citep{zadeh2016mosi}, and visual question answering~\citep{antol2015vqa}.
    For the second category that involves modality conversion, well-known examples are image captioning~(IC)~\citep{bernardi2016automatic}, where the task is to generate a textual description from an image, automatic speech recognition~(ASR)~\citep{yu2016automatic}, where the task is to transcribe spoken language audio into text, and speech synthesis~\citep{ling2015deep}, which is the converse of ASR, with the goal of generating speech from written language.
    
\begin{figure}[t]
        \centering
        \includegraphics[width=.7\textwidth]{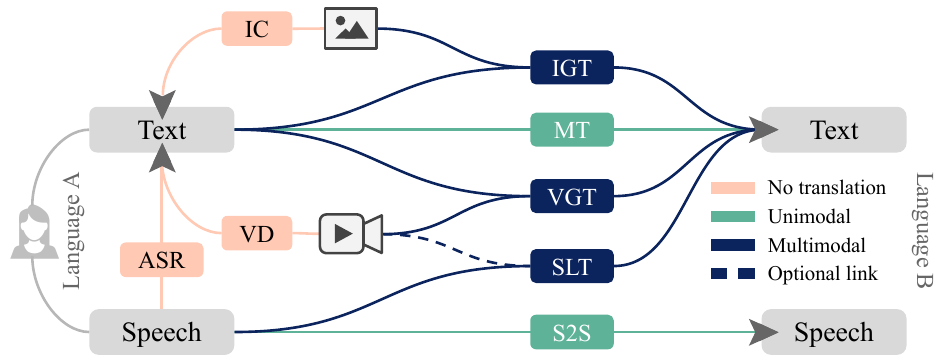}
        \caption{Prominent examples of multimodal translation tasks, such as image-guided translation~(IGT), video-guided translation~(VGT), and spoken language translation~(SLT), shown in contrast to unimodal translation tasks, such as text-based machine translation (MT) and speech-to-speech translation~(S2S), and multimodal NLP tasks that do not involve translation, such as automatic speech recognition~(ASR), image captioning~(IC), and video description~(VD).}
        \label{fig:tasks}
    \end{figure}

    Although more pointers exist in general surveys of multimodality in NLP~\citep{bernardi2016automatic, baltrusaitis-multimodal-2017,kafle2017visual,mogadala-trends-2019},
    this article is concerned with tasks that involve both multiple modalities and different input and output languages, \ie{}the tasks that fall under the umbrella of multimodal machine translation (MMT).
    The connection between modalities and translation tasks according to our definition is illustrated in Figure~\ref{fig:tasks},
    outlining the major tasks of spoken language translation~(SLT)~\citep{akiba-overview-2004}, image-guided translation~(IGT)~\citep{elliott2015multi,specia-shared-2016}, and video-guided translation~(VGT)~\citep{sanabria-how2:-2018,wang-vatex-2019}.
    
    Today, the rising interest in MMT is largely driven by the state-of-the-art performance and the architectural flexibility of neural sequence-to-sequence models~\citep{sutskever-sequence-2014,bahdanau-neural-2015,vaswani-attention-2017}. This flexibility, which is due to the end-to-end nature of these approaches, has the potential of bringing the vision, speech and language processing communities back together.
    From a historical point of view however,
    there was already a great deal of interest in doing machine translation~(MT) with non-text modalities, even before the arrival of successful statistical machine translation models. Among the earliest attempts is the Automatic Interpreting Telephony Research project~\citep{morimoto-automatic-1990}, a 1986 proposal that aimed at implementing a pipeline of automatic speech recognition, rule-based machine translation, and speech synthesis, making up a full speech-to-speech translation system. Further research has led to several other speech-to-speech translation systems~\citep{lavie-janus-iii:-1997,takezawa-japanese--english-1998,wahlster-mobile-2000}.
    
    In contrast, the use of visual modality in translation has not attracted comparable interest until recently. At present, there is a variety of multimodal task formulations including some form of machine translation,
    involving image captions, instructional text with photographs, video recordings of sign language, subtitles for videos~(and especially movies), and descriptions of video scenes. As a consequence, modern multimodal MT studies dealing with visual~(or audiovisual) information are becoming as prominent as those tackling audio. We believe that multimodal MT is a better reflection of how humans acquire and process language, with many theoretical advantages in language grounding over text-based MT as well as the potential for new practical applications like cross-modal cross-lingual information retrieval~\citep{gella2017image,kadar2018lessons}.

    In the following, we will provide a detailed description of MMT tasks and approaches that have been proposed in the past.
    Section~\ref{sec:tasks} contains an overview of the tasks of spoken language translation,
    image-guided translation and video-guided translation. Section~\ref{sec:eval} reviews the methods and caveats of evaluating MT performance, and discusses prominent evaluation campaigns, while Section~\ref{sec:datasets} contains an overview of major datasets that can be used as training or test corpora.
    Section~\ref{sec:models} discusses the state-of-the-art models and approaches in MMT, especially focusing on image-guided translation and spoken language translation.
    Section~\ref{sec:future} outlines fruitful directions of future research in multimodal MT.

\section{Tasks}
\label{sec:tasks}

    While our definition of multimodal machine translation excludes both cross-modal conversion tasks with no cross-linguality~(\eg{automatic speech recognition and video description}), and machine translation tasks within a single modality~(\eg{text-to-text and speech-to-speech translation}), it is still general enough to accommodate a fair variety of tasks. Some of these tasks such as spoken language translation~(SLT) and continuous sign language recognition~(CSLR) meet the criteria because their source and target languages are, by definition, expressed through different modes. Other tasks like image-guided translation~(IGT) and video-guided translation~(VGT) are included on the grounds that they complement the source language with related visuals that constitute an extra modality. In some cases, a well-established multimodal machine translation task can be characterised by methodological constraints~(\eg{simultaneous interpretation}), or by domain and semantics~(\eg{video description translation}).
    
    We observe that a shared modality composition is the foremost prerequisite that dictates the applicability of data, approaches and methodologies across multimodal translation tasks. For this reason, further in this article, we classify the studies we have surveyed according to the modality composition involved. We also restrict the scope of our discussions to the more well-recognised cases that involve audio and/or visual data in addition to text. In the following subsections, we explain our use of the terms \textit{spoken language translation}, \textit{image-guided translation}, and \textit{video-guided translation}, and provide further discussions for each of these tasks.

    

        
    \subsection{Spoken language translation}
    \label{sec:tasks:slt}
        
        Spoken language translation~(SLT), also known as speech-to-text translation or automatic speech translation, comprises the translation of speech in a source language to text in a target language. As such, it differs from conventional MT in the source-side modality. The need to simultaneously perform both modality conversion and translation means that systems must learn a complex input--output mapping, which poses a significant challenge. The SLT task has been shaped by a number of influential early works~\citep[\eg{}][]{vidal-finite-1997,ney-speech-1999}, and championed by the speech translation tasks of the IWSLT evaluation campaign since 2004~(see Section~\ref{sec:eval:tasks:iwslt}).
                    
        Traditionally, SLT was addressed by a pipeline approach~(see Section~\ref{sec:models} for more details), effectively separating multimodal MT into modality conversion followed by unimodal MT. More recently, end-to-end systems have been proposed, often based on NMT architectures, where the source language audio sequence is directly converted to the target language text sequence~\citep{weiss-s2s-st-2017, berard-e2e-audiobooks-2018}.
        Despite the short time during which end-to-end approaches have been developed, they have been rapidly closing the gap with the dominant paradigm of pipeline systems. The current state of end-to-end systems is discussed further in Section \ref{sec:slt:endtoend}.

    \subsection{Image-guided translation}
    \label{sec:tasks:ict}
    
    
        

        
        Image-guided translation can be defined as a contextual grounding task,
        where, given a set of images and associated documents, the aim is to enhance the translation of the documents by leveraging their semantic correspondence to the images. Resolving ambiguities through visual cues is one of the main motivating forces behind this task.
        
        A well-known realisation of IGT is image caption translation, where the correspondence is related to sentences being the descriptions of the images.
        Initial attempts at image caption translation were mostly pipeline approaches: \citet{elliott2015multi} proposed a pipeline of visually conditioned neural language models, while \citet{hitschler-2016-multimodal} approached the problem from a multimodal retrieval and reranking perspective. With the introduction of the WMT multimodal translation shared task~\citep[see Section~\ref{sec:eval:tasks:wmt}]{specia-shared-2016}, IGT attracted a lot more attention from the research community. Today, the prominent approaches rely on visually conditioning end-to-end neural MT systems with visual features extracted from state-of-the-art pretrained CNNs.
        
        Although the utility of the visual modality has recently been disputed under specific dataset and task conditions~\citep{elliott2018adversarial,caglayan-probing-2019}, using images when translating captions is theoretically very advantageous to handle grammatical characteristics~(\eg{noun genders}) in translating between dissimilar languages, and resolving translational ambiguities.
        Also, \citet{caglayan-probing-2019} shows how state-of-the-art models become capable of leveraging the visual signal
        when source captions are deliberately deteriorated in a simulated low-resource scenario. We discuss the current state of the art and the predominant approaches in IGT in Section~\ref{sec:ict}.

        

    
    \subsection{Video-guided translation}
    \label{sec:tasks:videosub}
        
        We posit the task of video-guided translation~(VGT) as a multimodal machine translation task similar to image-guided translation, but tackling video clips~(and potentially audio clips as well) rather than static images associated with the textual input. Within video-guided translation, there can be variants depending on the textual content. The source text can be transcripts of speech from the video, which would be typically segmented as standard subtitles, or a textual description of the visual scene or an action demonstrated in the clip, often created for visually impaired people. As such, video-guided translation can be subject to particular challenges from both SLT~(time-variant audiovisual input) and IGT~(indirect correspondence between source modalities). On the other hand, these similarities could also indicate that it might be possible to adapt or reuse approaches from both of those areas to bootstrap VGT systems.
        
        One major challenge hindering progress in video-guided translation is the relative scarcity of datasets. While a large collection such as the OpenSubtitles corpus\footnote{Derived from \url{https://www.opensubtitles.com/}}~\citep{lison-opensubtitles2016:-2016}
        can provide access to a considerable amount of parallel subtitles, there is no attached audiovisual content since the corresponding movies are not freely available. Recent efforts to compile freely accessible data for video-guided translation, like the How2~\citep{sanabria-how2:-2018} and \vatex{}~\citep{wang-vatex-2019} datasets~(both described in Section~\ref{sec:datasets:vgt}) have started to alleviate this bottleneck. Although there has been decidedly little time to observe the full impact of such initiatives, we hope that they will inspire further research in video-guided translation.
    


\section{Evaluation}
\label{sec:eval}
    
    Evaluating the performance of a machine translation system is a difficult and controversial problem. Typically, there are numerous ways of translating even a single sentence which would be acceptably produced by human translators~(or systems), and it is often unclear which one is~(or which ones are) good or better, and in what respect, given that the pertinent evaluation criteria are multi-dimensional, context-dependent, and highly subjective~\citep[see for example][]{chesterman-can-2002,drugan-quality-2013}. Traditionally, human analysis of translation quality has often been divided into the evaluation of adequacy~(semantic transfer from source language) and fluency~(grammatical soundness of target language)~\citep{doherty-human-2017}. While this separation is considered somewhat artificial, it was created to make evaluation simpler and 
    to allow comparison of translation systems in more specific terms. In practice, systems that are good at one criterion tend to be good at the other, and a lot of the more recent evaluation campaigns have focused on directly ranking systems for general quality rather than scoring individual systems on these criteria~(relative ranking), or scoring systems for general quality instead~(direct assessment). 
    
    Since human evaluation comes with considerable monetary and time costs~\citep{castilho-approaches-2018}, evaluation efforts have converged to devising automatic metrics in recent years~\citep{ma-results-2018,ma-results-2019}, which typically operate by comparing the output of a translation system against one or more human translations.
    While a number of metrics have been proposed over the last two decades, they are mostly based on statistics computed between the translation hypothesis and one or more references.
    Procuring reference translations in itself entails some costs, and any metrics and approaches that require multiple references to work well may therefore not be feasible for common use.
    Further in this section, we discuss the details of some of the dominant evaluation metrics as well as the most well-known shared tasks of multimodal MT that serve as standard evaluation settings to facilitate research.
    
    
    \subsection{Metrics}
    \label{sec:eval:metrics}
        
        Among the various MT evaluation metrics in the literature, the most commonly used ones are  BLEU~\citep{papineni-bleu:-2001}, METEOR~\citep{lavie-meteor:-2007,denkowski-meteor-2014} and TER~\citep{snover-study-2006}. To summarise them briefly, BLEU is based on an aggregate precision measure of n-gram matches between the reference(s) and machine translation, and penalises translations that are too short. METEOR accounts for and gives partial credit to stem, synonyms, and paraphrase matches, and considers both precision and recall with configurable weights for both criteria.  TER is a variant of word-level edit distance between the source and the target sentences, with an added operation for shifting one or more adjacent words. 
        BLEU is by far the most commonly used automatic evaluation metric, despite its relative simplicity.
        Most quantitative comparisons of machine translation systems are reported using only BLEU scores. METEOR has been shown to correlate better with human judgements (especially for adequacy) due to both its flexibility in string matching and its better balance between precision and recall, but its dependency on linguistic resources makes it less applicable in the general case. Both BLEU and METEOR, much like the majority of other evaluation metrics developed so far, are reference-based metrics. These metrics are inadvertently heavily biased on the translation styles that they see in the reference data, and end up penalising any alternative phrasing that might be equally correct~\citep{fomicheva-specia_ACL:2016}.
        
        Human evaluation is the optimal choice when a trustworthy measure of translation quality is needed and resources to perform it are available. The usual strategies for human evaluation are fluency and adequacy rankings, direct assessment~(DA)~\citep{graham-etal-2013-continuous}, and post-editing evaluation~(PE)~\citep{snover-study-2006}. Fluency and adequacy rankings are conventionally between 1--5, while DA is a general scale between 0--100 indicating how ``good'' the translation is, either with respect the original sentence in the source language (DA-\emph{src}), or the ground truth translation in the target language (DA-\emph{ref}).
        On the other hand, in PE, human annotators are asked to \emph{correct} translations by changing the words and the ordering as little as possible, and the rest of the evaluation is based on an automatic edit distance measure between the original and post-edited translations, or other metrics such as post-editing time and keystrokes~\citep{specia-etal_MTSummit:2017}. For pragmatics reasons, these human evaluation methods are typically crowdsourced to non-expert annotators to reduce costs. While this may still result in consistent evaluation scores if multiple crowd annotators are considered, it is a well-accepted fact that professional translators capture more details and are generally better judges than non-expert speakers~\citep{bentivogli-machine-2018}.
        
        The problems recognised even in human evaluation methods substantiate the notion that no metric is perfect. In fact, evaluation methods are an active research subject in their own right~\citep{specia-findings-2018,ma-results-2018,ma-results-2019}. However, there is currently little research on developing evaluation approaches specifically tailored to multimodal translation. Fully-automatic evaluation is typically text-based, while methods that go beyond the text rely on manually annotated resources, and could rather be considered semi-automatic.
        One such method is multimodal lexical translation~(MLT)~\citep{lala-multimodal-2018}, which is a measure of translation accuracy for a set of ambiguous words given their textual context and an associated image that allows visual disambiguation.
        %
        %
        Even in human evaluation there are only a few examples where the evaluation is multimodal, such as the addition of images in the evaluation of image caption translations via direct assessment~\citep{elliott-findings-2017, barrault-findings-2018}, or via qualitative comparisons of post-editing~\citep{frank-assessing-2018}. Having consistent methods to evaluate how well translation systems take multimodal data into account would make it possible to identify bottlenecks and facilitate future development. One possible promising direction is the work of \cite{madhyastha-etal-2019-vifidel} for image captioning evaluation, where the content of the image is directly taken into account via the matching of detected objects in the image and concepts in the generated caption.
    
        
    \subsection{Shared tasks}
    \label{sec:eval:tasks}
    
        A great deal of research into developing natural language processing systems is made in preparation for shared tasks under academic conferences and workshops, and the relatively new subject of multimodal machine translation is not an exception. These shared tasks lay out a specific experimental setting for which participants submit their own systems, often developed using the training data provided by the campaign. Currently, there are not many datasets encompassing both multiple languages and multiple modalities that are also of sufficiently high quality and large size, and available for research purposes. However, multilingual datasets that augment text with only speech or only images are somewhat less rare than those with videos, given their utility for tasks such as automatic speech recognition and image captioning. Adding parallel text data in other languages enables such datasets to be used for spoken language translation and image-guided translation, both of which are represented in shared tasks organised by the machine translation community. The Conference on Machine Translation (WMT) ran three shared tasks for image caption translation from 2016--2018, and the International Workshop on Spoken Language Translation (IWSLT) has led an annual evaluation campaign on speech translation since 2004.
            
        
        \subsubsection{Image-guided translation: WMT multimodal translation task}
        \label{sec:eval:tasks:wmt}
        
            The Conference on Machine Translation (WMT) has organised multimodal translation shared tasks annually since the first event~\citep{specia-shared-2016} in 2016.
            The first shared task was such that the participants were given images and an English caption for each image as input, and were required to generate a translated caption in German. The second shared task had a similar experimental setup, but added French to the list of target languages, and new test sets. The third shared task in 2018 added Czech as a third possible target language, and another new test set. This last\footnote{The multimodal translation task was not held in WMT 2019.} task also had a secondary track which only had Czech on the target side, but allowed the use of English, French and German captions together along with the image in a multisource translation setting.
            
            The WMT multimodal translation shared tasks evaluate the performances of submitted systems on several test sets at once, including the Ambiguous COCO test set~\citep{elliott-findings-2017}, which incorporates image captions that contain ambiguous verbs~(see Section~\ref{sec:datasets:mscoco}). The translations generated by the submitted systems are scored by the METEOR, BLEU, and TER metrics. In addition, all participants are required to devote resources to manually scoring translations in a blind fashion. This scoring is done by direct assessment using the original source captions and the image as references. 
            During the assessment, ground truth translations are shuffled into the outputs from the submissions, and scored just like them. This establishes an approximate reference score for the ground truth, and the submitted systems are analysed in relation to this.
        
        
        \subsubsection{Spoken language translation: IWSLT evaluation campaign}
        \label{sec:eval:tasks:iwslt}
            
            The spoken language translation tasks have been held as part of the annual IWSLT evaluation campaign since
            \citet{akiba-overview-2004}.
            Following the earlier C-STAR evaluations, the
            aim of the campaign is to investigate
            newly-developing translation technologies as well as methodologies for evaluating them.
            The first years of the campaign were based on a basic travel expression corpus developed by C-STAR to facilitate standard evaluation, containing basic tourist utterances (\eg{\emph{``Where is the restroom?''}}) and their transcripts. The corpus was eventually extended with more samples (from a few thousand to tens of thousands) and more languages (from Japanese and English, to Arabic, Chinese, French, German, Italian, Korean, and Turkish). Each year also had a new challenge theme, such as robustness of spoken language translation, spontaneous (as opposed to scripted) speech, and dialogue translation, introducing corresponding data sections (\eg{running dialogues}) as well as sub-tasks (\eg{translating from noisy ASR output}) to facilitate the challenges. Starting with \citet{paul-overview-2010}, the campaign adopted TED talks as their primary training data, and eventually shifted away from the tourism domain towards lecture transcripts.
            
            Until \citet{cettolo-iwslt-2016}, the evaluation campaign had three main tracks: Automatic speech recognition, text-based machine translation, and spoken language translation. While these tasks involve different sources and diverging methodologies, they converge on text output. The organisers have made considerable effort to use several automatic metrics at once to evaluate participating systems, and to analyse the outputs from these metrics. Traditionally, there has also been human evaluation 
            on the most successful systems for each track according to the automatic metrics. These assessments have been used to investigate which automatic metrics correlate with which human assessments to what extent, and to pick out and discuss drawbacks in evaluation methodologies.
            
            Additional tasks such as 
            dialogue translation~\citep{cettolo-overview-2017} and low-resource spoken language\linebreak translation~\citep{niehues-iwslt-2018} were reintroduced to the IWSLT evaluation campaign from 2017 on, as TED data and machine translation literature both grew richer. \citet{niehues-iwslt-2019} introduced a new audiovisual spoken language translation task, leveraging the How2 corpus~\citep{sanabria-how2:-2018}. In this task, video is included as an additional input modality, for the general case of subtitling audiovisual content.


\section{Datasets}
\label{sec:datasets}
    Text-based machine translation has recently enjoyed widespread success with the adoption of deep learning model architectures. The success of these data-driven systems rely heavily on the factor of data availability. An implication of this for multimodal MT is the need for large datasets in order to keep up with the data-driven state-of-the-art methodologies. Unfortunately, due to its simultaneous requirement of multimodality and multilinguality in data, multimodal MT is subject to an especially restrictive bottleneck. Datasets that are sufficiently large for training multimodal MT models are only available for a handful of languages and domain-specific tasks. The limitations imposed by this are increasingly well-recognised, as evidenced by the fact that most major datasets intended for multimodal MT were released relatively recently. Some of these datasets are outlined in Table~\ref{table:datasets:summary}, and explained in more detail in the subsections to follow.

    
    \subsection{Image-guided translation datasets}
    \label{sec:datasets:ict}
        
\begin{figure}[t]
            \centering
            \includegraphics[width=.9\textwidth]{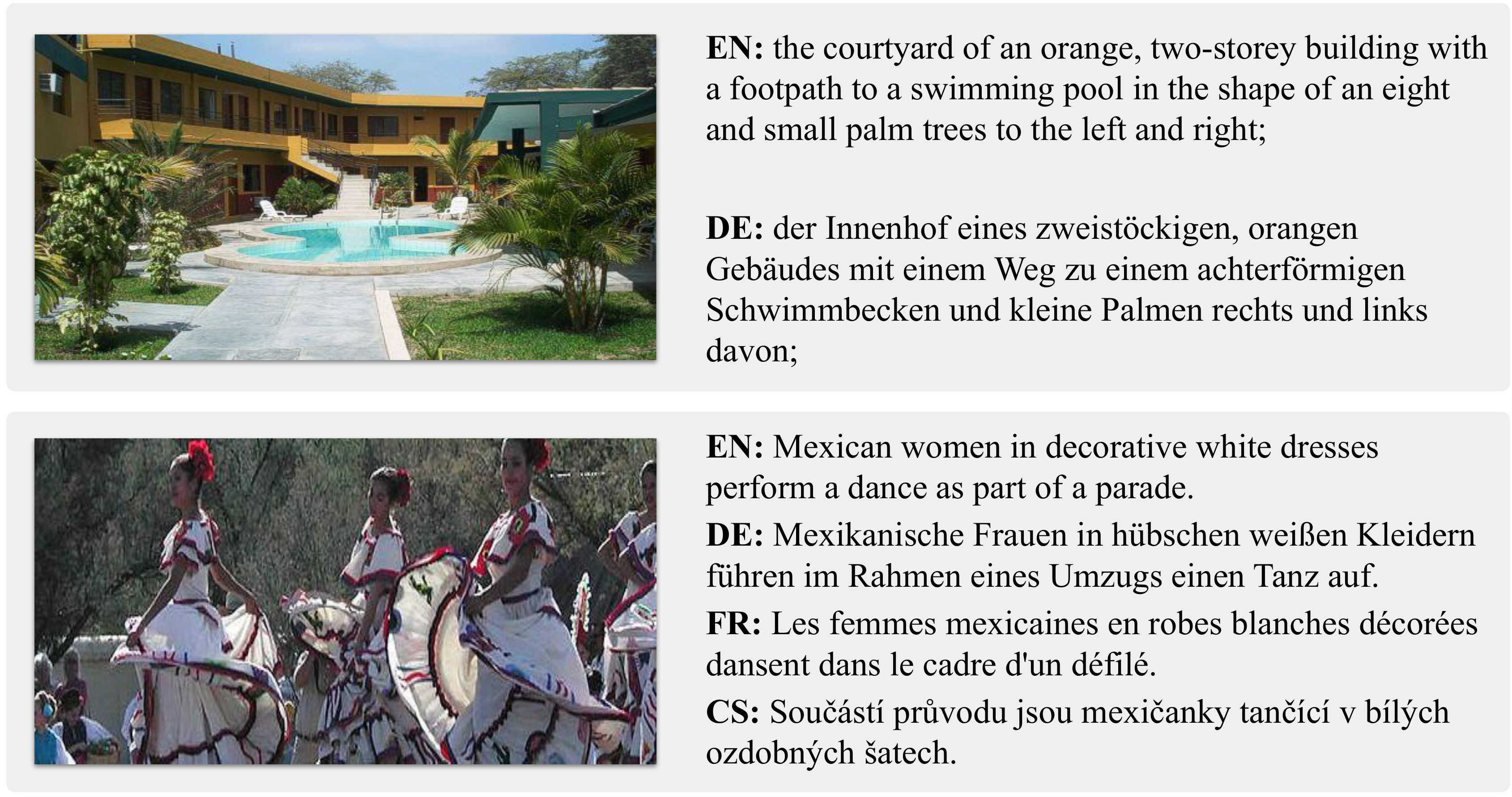}
            \caption{Contrasting examples from IAPR TC-12 image descriptions (top) and Multi30k image captions (bottom).}
            \label{fig:datasets:igt}
        \end{figure}

        
        \element{IAPR TC-12}
            The International Association of Pattern Recognition (IAPR) TC-12 benchmark dataset~\citep{grubinger-iapr-2006} was created for the cross-language image retrieval track of the CLEF evaluation campaign (ImageCLEF 2006)~\citep{clough-overview-2006}. The benchmark is structurally similar to the multilingual image caption datasets commonly used by contemporary image-guided translation systems. IAPR TC-12 contains 20,000 images from a collection of photos of landmarks taken in various countries, provided by a travel organisation. Each image was originally annotated with German descriptions, and later translated to English. These descriptions are composed of phrases that describe the visual contents of the photo following strict linguistic patterns, as shown in Figure~\ref{fig:datasets:igt}. The dataset also contains \textit{light} annotations such as titles and locations in English, German, and Spanish. \\
        
        
        
        \element{Flickr8k}
            Released in 2010, the Flickr8k dataset~\citep{rashtchian-collecting-2010} has been one of the most widely-used multimodal corpora. Originally intended as a high-quality training corpus for automatic image captioning, the dataset comprises a set of 8,092 images extracted from the Flickr website, each with 5 crowdsourced captions in English that describe the image. Flickr8k has shorter captions compared to IAPR TC-12, focusing on the most salient objects or actions, rather than complete descriptions. As the dataset has been a popular and useful resource, it has been further extended with captions in other languages such as Chinese~\citep{li-adding-2016} and Turkish~\citep{unal-tasviret:-2016}. However, as these captions were independently crowdsourced, they are not translations of each other, which makes them less effective for MMT. \\
        
        
        
        \element{Flickr30k / Multi30k}
            The Flickr30k dataset~\citep{young-image-2014} was released in 2014 as a larger dataset following in the footsteps of Flickr8k. Collected using the same crowdsourcing approach for independent captions as its predecessor, Flickr30k contains 31,783 photos depicting common scenes, events, and actions, each annotated with 5 independent English captions. 
            Multi30k~\citep{elliott-multi30k:-2016} was initially released as a bilingual subset of Flickr30k captions, providing German translations for 1 out of the 5 English captions per image, with the aim of stimulating multimodal and multilingual research. In addition, the study collected 5 independent German captions for each image. The WMT multimodal translation tasks later introduced French~\citep{elliott-findings-2017} and Czech~\citep{barrault-findings-2018} extensions to Multi30k, making it a staple dataset for image-guided translation, and further expanding the set's utility to cutting-edge subtasks such as multisource training. An example from this dataset can be seen in Figure~\ref{fig:datasets:igt}. \\
        
        \begin{table}[t]
        \centering
        \caption{Summary statistics from most prominent multimodal machine translation datasets. We report image captions \textit{per language}, and audio clips and segments \textit{per language pair}.}
        \label{table:datasets:summary}
        \renewcommand{\arraystretch}{1.17}
        \begin{tabular}{@{}llllcccc@{}}
            \toprule
            
            \textbf{Dataset} & \textbf{Media} & \textbf{Text} & \textbf{Languages} & \taskf{SLT} & \taskf{IGT} & \taskf{VGT} \\
            
            \midrule
            
            IAPR TC-12~\citep{grubinger-iapr-2006}       & 20k images    & 20k captions        & \textsc{de,\,en}           & & \tick & \\
            Flickr8k~\citep{rashtchian-collecting-2010}  & 8k images     & 41k captions       & \textsc{en,\,tr,\,zh}      & & \tick & \\
            Flickr30k~\citep{young-image-2014}           & 30k images    & 158k captions      & \textsc{de,\,en}           & & \tick & \\
            Multi30k~\citep{elliott-multi30k:-2016}      & 30k images    & 30k captions       & \textsc{cs,\,de,\,en,\,fr} & & \tick & \\
            
            \midrule
            
            
            
            QED~\citep{abdelali-amara-2014}  & 23.1k video clips & 8k--335k segments & 20 languages     & \tick & & \tick \\
            How2~\citep{sanabria-how2:-2018} & 13k video clips   & 189k segments     & \textsc{en,\,pt} & \tick & & \tick \\
            \vatex{}~\citep{wang-vatex-2019} & 41k video clips   & 206k segments     & \textsc{en,\,zh} &       & & \tick \\
            
            \midrule
            
            
            
            WIT$^3$~\citep{cettolo-wit3:-2012}                     & 2,086 audio clips   & 3--575k segments    & 109 languages                   & \tick & & \\
            Fisher \& Callhome~\citep{post-improved-2013}          & 38h audio      & 171k segments       & \textsc{en,\,es}                & \tick & & \\
            MSLT~\citep{federmann-microsoft-2017}                  & 4.5--10h audio & 7k--18k segments    & \textsc{de,\,en,\,fr,\,ja,\,zh} & \tick & & \\
            \tedtrans{}~\citep{niehues-iwslt-2018}                 & 1,565 audio clips   & 171k segments       & \textsc{de,\,en}                & \tick & & \\
            LibriSpeech~\citep{kocabiyikoglu-librispeech-slt-2018} & 236h audio     & 131k segments       & \textsc{en,\,fr}                & \tick & & \\
            MuST-C~\citep{gangi-mustc-2019}                        & 385--504h audio    & 211k--280k segments & 10 languages                    & \tick & & \\
            MaSS~\citep{boito-mass-2019}                           & 18.5--23h audio     & 8.2k segments       & 8 languages                     & \tick & & \\
            
            
            \bottomrule
        \end{tabular}
    \end{table}

            
            
            
            
        
        
        \element{WMT test sets}
            The past three years of multimodal shared tasks at WMT each came with a designated test set for the task ~\citep{specia-shared-2016, elliott-findings-2017, barrault-findings-2018}. Totalling 3,017 images in the same domain as the Flickr sets~(including Multi30k), these sets are too small to be used for training purposes, but could smoothly blend in with the other Flickr sets to expand their size. So far, test sets from the previous shared tasks (each containing roughly 1,000 images with captions) have been allowed for validation and internal evaluation. In parallel with the language expansion of Multi30k, the test set from 2016 contains only English and German captions, and the one from 2017 contains only English, German, and French. The 2018 test set contains English, German, French, and Czech captions that are not publicly available, though systems can be evaluated against it using an online server.\footnote{\url{https://competitions.codalab.org/competitions/19917}}\\
        
        
        
        
        \label{sec:datasets:mscoco}            
        \element{MS COCO Captions}
            Introduced in 2015, the MS COCO Captions dataset~\citep{chen-microsoft-2015} offers caption annotations for a subset of roughly 123,000 images from the large-scale object detection and segmentation training corpus MS COCO (Microsoft Common Objects in Context)~\citep{fleet-microsoft-2014}. Each image in this dataset is associated with up to 5 independently annotated English captions, with a total of 616,767 captions. Though originally a monolingual dataset, the dataset's large size makes it useful for data augmentation methods for image-guided translation, as demonstrated in~\citet{gronroos-memad-2018}. There has also been some effort to add other languages to COCO. A small subset with only 461 captions containing ambiguous verbs was released as a test set for the WMT 2017 multimodal machine translation shared task, called Ambiguous COCO~\citep{elliott-findings-2017}, and is available in all target languages of the task. The YJ Captions dataset~\citep{miyazaki-cross-2016} and the STAIR Captions dataset~\citep{yoshikawa-stair-2017} comprise, respectively, 132k and 820k crowdsourced Japanese captions for COCO images. However, these are not parallel to the original English captions, as they were independently annotated.
            
        
    \subsection{Spoken language translation datasets}
    \label{sec:datasets:ted}
        
        
        \element{The TED corpus}
            TED is a nonprofit organisation that hosts talks in various topics, comprising a rich resource of spoken language produced by a variety of speakers in English. Video recordings of all TED talks are made available through the TED website\footnote{\url{http://www.ted.com/talks}}, as well as transcripts with translations in up to 116 languages. While the talks comprise a rich resource for language processing, the original transcripts are divided into arbitrary segments formatted like subtitles, which makes it difficult to get an accurate sentence-level parallel segmentation for use in translation systems. While resegmentation is possible with heuristic approaches, it comes with the additional challenge of aligning the new segments to the audiovisual content, and to each other in source and target languages.
            The Web Inventory of Transcribed and Translated Talks (WIT$^3$)~\citep{cettolo-wit3:-2012} is a resource with the aim of facilitating the use of the TED Corpus in MT. The initiative distributes transcripts organised in XML files through their website\footnote{\url{http://wit3.fbk.eu}}, as well as tools to process them in order to extract parallel sentences. Currently, WIT$^3$ covers 2,086 talks in 109 languages containing anywhere between 3 and 575k segments in raw transcripts, and is continually growing.
            
            Since 2011, the annual speech translation tracks of the IWSLT evaluation campaign~(see Section~\ref{sec:eval:tasks:iwslt}) has used datasets compiled from WIT$^3$. While each of these sets contain a high-quality selection of English transcripts aligned with the audio and the target languages featured each year, they are not useful for training SLT systems due to their small sizes. As part of the 2018 campaign, the organisers released a large-scale English--German corpus~\citep{niehues-iwslt-2018} containing 1,565 talks with 170,965 segments automatically aligned based on time overlap, which allows end-to-end training of SLT models. 
            The MuST-C dataset~\citep{gangi-mustc-2019} is a more recent effort to compile a massively multilingual dataset from TED data, spanning 10 languages (English aligned with Czech, Dutch, French, German, Italian, Portuguese, Romanian, Russian, and Spanish translations), using more reliable timestamps for alignments than the \tedtrans{} dataset using a rigorous alignment process.
            The dataset contains a large amount of data for each target language, corresponding to a selection of English speech ranging from 385 hours for Portuguese to 504 hours for Spanish. \\
            
        
        \element{LibriSpeech}
            The original LibriSpeech corpus \citep{panayotov-librispeech-2015} is a collection of 982 hours of read English speech derived from audiobooks from the LibriVox project, automatically aligned to their text versions available from the Gutenberg project for the purpose of training ASR systems.
            \citet{kocabiyikoglu-librispeech-slt-2018} augments this dataset for use in training SLT systems by aligning chapters from LibriSpeech with their French equivalents through a multi-stage automatic alignment process.
            The result is a parallel corpus of spoken English to textual French,
            consisting of 1408 chapters from 247 books,
            totalling 236 hours of English speech and approximately 131k text segments.\\

        
        \element{MSLT}
            The Microsoft Speech Language Translation~(MSLT) corpus~\citep{federmann-microsoft-2016} consists of bilingual conversations on Skype,
            together with transcriptions and translations.
            For each bilingual speaker pair, there is one conversation where the first speaker uses their native language and the second speaker uses English,
            and another with the roles reversed.
            The first phase transcripts were annotated for disfluencies, noise and code switching.
            In a second phase, the transcripts were cleaned, punctuated and recased.
            The corpus contains 7 to 8 hours of speech for each of English, German, and French.
            The English speech was translated to both German and French,
            while German and French speech was translated only to English.
            \citet{federmann-microsoft-2017} repeat the process with Japanese and Chinese, expanding the dataset with 10 hours of Japanese and 4.5 hours of Chinese speech. \\

        
        
        
        \element{Fisher \& Callhome}
            \citet{post-improved-2013} extends the Fisher\footnote{Speech: \url{https://catalog.ldc.upenn.edu/LDC2010S01}, Transcripts: \url{https://catalog.ldc.upenn.edu/LDC2010T04}} and Callhome\footnote{Speech: \url{https://catalog.ldc.upenn.edu/LDC96S35}, Transcripts: \url{https://catalog.ldc.upenn.edu/LDC2010T04}} datasets of transcribed Spanish speech with English translations, developed by the Linguistic Data Consortium. The original Fisher dataset contains about 160 hours of telephone conversations in various dialects of Spanish between strangers, while the Callhome dataset contains 20 hours of telephone conversations between relatives and friends. The translations were collected from non-professional translators on the crowdsourcing platform Mechanical Turk. Fisher~\&~Callhome
            is distributed with predesignated development and test splits, a part of which contains four reference translations for each transcript segment.
            The data in the corpus also includes ground truth ASR lattices that facilitate the training of strong specialized ASR models, allowing pipeline SLT studies to focus on the MT component.
            As the largest SLT corpus available at the time of its release, the Fisher~\&~Callhome corpus has been widely used, and remains relevant for SLT today. \\ 

        
        \element{MaSS}
            The \textbf{M}ultilingu\textbf{a}l corpus of \textbf{S}entence-aligned \textbf{S}poken utterances (MaSS) \citep{boito-mass-2019} is a multilingual corpus of read bible verses and chapter names from the New Testament.
            It is fully multi-parallel across 8 languages~(Basque, English, Finnish, French, Hungarian, Romanian, Russian, and Spanish), comprising 56 language pairs in total.
            The multi-parallel content makes this dataset suitable for training SLT systems for language pairs not including English, unlike other multilingual datasets such as MuST-C.
            The data is aligned on the level of verses, rather than sentences. In rare cases, the audio for some verses is missing for some languages.
            MaSS contains a total of 8,130 eight-way parallel text segments, corresponding to anywhere between 18.5 and 23 hours of speech per language.
    
    
        
        
        
        
        
        
    
    
    \subsection{Video-guided translation datasets}
    \label{sec:datasets:vgt}
    
        \element{The QED corpus}
            The QCRI Educational Domain (QED) Corpus~\citep{guzman-amara-2013,abdelali-amara-2014}, formerly known as the QCRI AMARA Corpus, is a large-scale collection of multilingual video subtitles. The corpus contains publicly available videos scraped from massive online open courses (MOOCs), spanning a wide range of subjects. The latest v1.4 release comprises a selection of 23.1k videos in 20 languages (Arabic, Bulgarian, Traditional and Simplified Chinese, Czech, Danish, Dutch, English, French, German, Hindi, Italian, Japanese, Korean, Polish, Portuguese, Russian, Spanish, Thai, and Turkish), subtitled in the collaborative Amara environment\footnote{\url{https://amara.org/}}~\citep{jansen-amara-2014} by volunteers. A sizeable portion of the videos has parallel subtitles in multiple languages, varying in size from 8k segments (for Hindi--Russian) to 335k segments (for English--Spanish). Of these, about 75\% of the parallel segments align perfectly in the original data, while the rest were automatically aligned using heuristic algorithms. An alpha v2.0 of the QED corpus is currently underway, scheduled to appear in the OPUS repository~\citep{tiedemann-parallel-2016}, containing a large amount of~(noisy) re-crawled subtitles. \\

\begin{figure}[t]
            \centering
            \includegraphics[width=.99\textwidth]{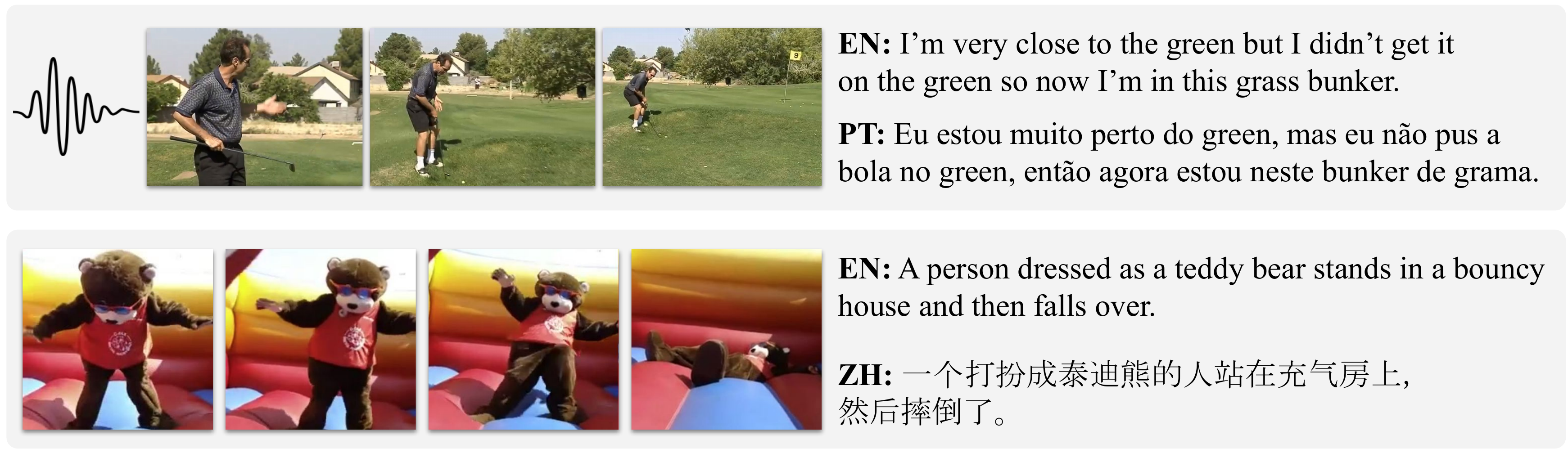}
            \caption{Examples from How2 video subtitles (top) and \vatex{} video descriptions (bottom),\\ retrieved and adapted from~\citet{sanabria-how2:-2018} and~\citet{wang-vatex-2019}, respectively.}
            \label{fig:datasets:vgt}
        \end{figure}

 
        \element{The How2 dataset}
            The How2 dataset~\citep{sanabria-how2:-2018} is a collection of 79,114 clips with an average length of 90 seconds, containing around 2,000 hours of instructional YouTube videos in English, spanning a variety of topics. The dataset is intended as a resource for several multimodal tasks, such as multimodal ASR, multimodal summarisation, spoken language translation, and video-guided translation. To establish cross-modal associations, the videos in the dataset were annotated with word-level alignments to ground truth English subtitles. There are also English descriptions of each video written by the users who uploaded the videos, added to the dataset as metadata corresponding to video-level summaries. For the purpose of multimodal translation, a 300-hours subset of How2 that covers 22 different topics is available with crowdsourced Portuguese translations. This dataset has also recently been used for multimodal machine translation~\citep{sanabria-how2:-2018, wu2019}. An example from this dataset can be seen in Figure~\ref{fig:datasets:vgt}. \\

        \element{The \vatex{} dataset}
            The \textbf{V}ideo \textbf{a}nd \textbf{TeX}t~(\vatex{}) dataset~\citep{wang-vatex-2019} is a bilingual collection of video descriptions, built on a subset of 41,250 video clips from the action classification benchmark DeepMind Kinetics-600~\citep{kay-kinetics-2017,carreira-short-2018}. Each clip runs for about 10 seconds, showing one of 600 human activities. \vatex{} adds 10 Chinese and 10 English crowdsourced captions describing each video, half of which are independent annotations, and the other half Chinese--English parallel sentences. With low-approval samples removed, the released version of the dataset contains 206,345 translation pairs in total. \vatex{} is intended to facilitate research in multilingual video captioning and video-guided machine translation, and the authors keep a blind test set reserved for use in evaluation campaigns. The rest of the dataset is divided into training (26k videos), validation (3k videos), and public test splits (6k videos). The training and validation splits also have public action labels. An example from \vatex{} is shown in Figure~\ref{fig:datasets:vgt}.


\section{Models and Approaches}
\label{sec:models}
    
    This section discusses the state-of-the-art models 
proposed to solve the multimodal machine translation (MMT) tasks introduced in Section~\ref{sec:tasks}.
For some MMT tasks, the traditional approach is to put together a pipeline to divide the task into several sub-tasks, and cascade different modules to handle each of them.
For instance, in the case of spoken language translation (SLT), this pipeline would first convert the input speech into text by an automatic speech recognition module (modality conversion), and then redirect the output to a text-based MT module. This is in contrast to end-to-end models, where the source language would be encoded into an intermediate representation, and decoded directly into the target language. Pipeline systems are less vulnerable to training data insufficiency compared to data-driven end-to-end systems, since each component can be pretrained in isolation on abundant sub-task resources. However, they carry the risk of error propagation between stages and ignore cross-modal transfer of implicit semantics. As an example for the latter, consider two languages which emphasise words via prosody and specific word order, respectively. Translating the transcript would make it impossible to reflect the word order in the target sentence as the semantic correspondence would be lost at transcription stage.
Nevertheless, both pipeline and end-to-end approaches rely heavily on the sequence-to-sequence learning framework on account of its flexibility and good performance across tasks. In the following, we describe this framework in detail.

\begin{figure}[t]
    \centering
    \includegraphics[width=.89\textwidth]{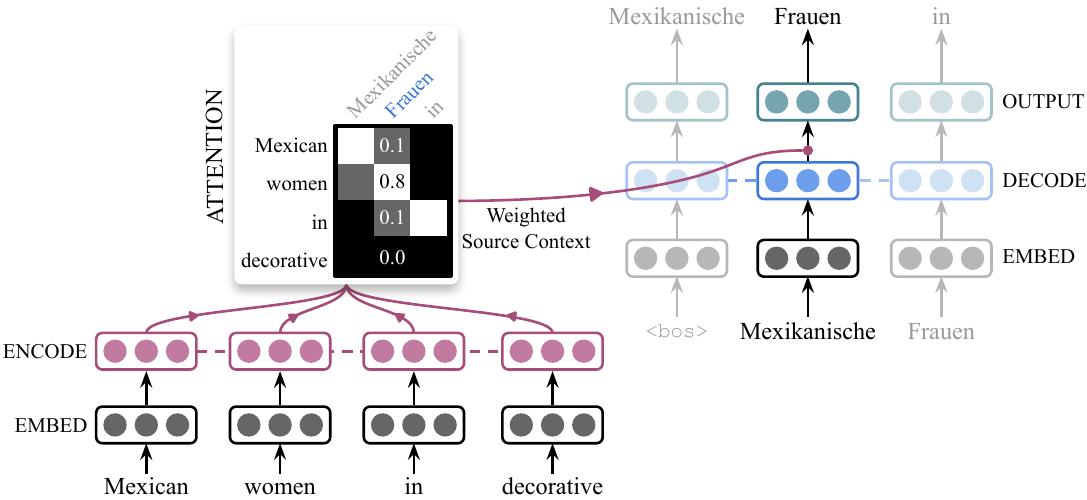}
    \caption{A simplified view of encoder-decoder architecture with attention: an English sentence is first encoded into a latent space from which an attentive decoder sequentially generates the German sentence. The dashed \emph{recurrent} connections are replaced by self-attention in fully-connected architectures such as transformers~\citep{vaswani-attention-2017}.}
    \label{fig:unimodal-mt:encdec}
\end{figure}

General purpose sequence-to-sequence learning is inspired by the pioneering works in unimodal neural machine translation (NMT). The state of the art in unimodal MT has been dominated by statistical machine translation (SMT) methodologies~\citep{koehn-smt-2009} for at least two decades, until the field drastically moved towards NMT techniques around 2015. Inspired by the successful use of deep neural networks in language modelling~\citep{bengio-nplm-2003,mikolov-rnnlm-2010} and automatic speech recognition~\citep{graves-speech-2013}, there has been a plethora of NMT studies featuring different neural architectures and learning methods. These architectures
often rely on continuous word vector representations to encode various kinds of linguistic information in a common vector space, thereby eliminating the need for
hand-crafted linguistic features.
One of the first NMT studies by~\citet{kalchbrenner-recurrent-2013} combined recurrent language modelling~\citep{mikolov-rnnlm-2010} and convolutional neural networks (CNN) to
improve the performance of SMT systems through rescoring.
Later on, the application of recurrent architectures, such as bidirectional RNNs~\citep{schuster-bidirectional-1997}, LSTMs~\citep{hochreiter-long-1997, graves-framewise-2005}, and GRUs~\citep{chung-empirical-2014}, introduced further diversity into the field, eventually leading to the fundamental encoder-decoder architecture~\citep{cho-properties-2014,sutskever-sequence-2014}.
These more advanced neural units were not as susceptible to the problems initially perceived in NMT, dealing naturally with variable-length sequences, and having clear computational advantages as well as superior performance. However, the difficulty of learning long-range dependencies in translation sequences~(\eg{grammatical agreement in very long sentences})
remained an issue until the introduction of the attention mechanism~\citep{bahdanau-neural-2015}. The attention mechanism addressed this issue by simultaneously learning to align translation units and to translate, supplying a context window with the relevant input units at each decoding step, \ie{}for each generated word in the target language~(Figure~\ref{fig:unimodal-mt:encdec}).
The performance of the NMT systems that followed came close to, and soon surpassed, that of the state-of-the-art SMT systems. Successful non-recurrent alternatives have also been proposed, such as convolutional encoders and decoders with attention~\citep{gehring-convolutional-2017}, and the fully-connected deep transformers which employ the idea of self-attention in addition to the default cross-attention mechanism~\citep{vaswani-attention-2017}. The main motivation behind these is to allow for efficient parallel training across multiple processing units, and to prevent learning difficulties such as vanishing gradients. 

Lastly, we would like to mention some major open-source toolkits which contribute vastly to the state of the art in machine translation by allowing fast prototyping of new approaches as well as the extension of existing ones to new tasks and paradigms:
Moses~\citep{koehn-moses:-2007} for SMT, and FairSeq~\citep{ott-fairseq-2019}, Lingvo~\citep{shen-lingvo-2019}, Marian~\citep{junczys-dowmunt-marian:-2018}, Nematus~\citep{sennrich-nematus-2017}, NeuralMonkey~\citep{helcl-neuralmonkey-2018}, nmtpytorch~\citep{caglayan-nmtpy-2017}, OpenNMT~\citep{klein-opennmt:-2017}, Sockeye~\citep{hieber-sockeye:-2017} and Tensor2Tensor~\citep{vaswani-tensor2tensor-2018} for NMT.
    \subsection{Image-guided translation}
    \label{sec:ict}
In this section, we present the state-of-the-art models for the image-guided translation~(IGT) task. We first discuss the visual feature extraction process, continue with reviews of the two main end-to-end neural approaches, and finally briefly cover retrieval and reranking methods.

\subsubsection{Feature extraction}
\label{sec:tasks:ict:feat}
\begin{figure}[t]
    \centering
    \includegraphics[width=.6\textwidth]{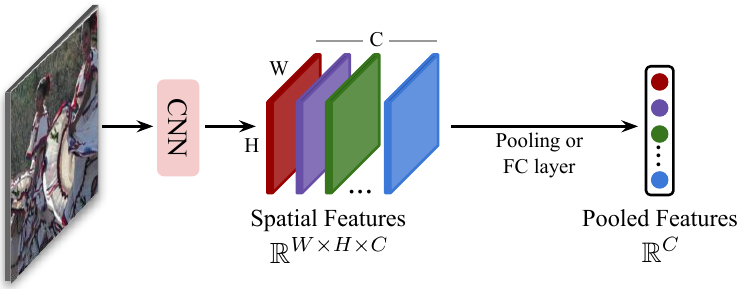}
    \caption{An overview of two common types of visual featuers extracted from CNNs.}
    \label{fig:ict:cnnfeats}
\end{figure}

The practice of embedding translation units into continuous vector representations has become a standard in NMT. For compatibility with various NMT architectures, multimodal MT systems need to embed input data from other modalities, whether alongside or in place of the text, in a similar fashion. For visual information, the current best practice is to use a convolutional neural network (CNN) with multiple layers stacked on top of each other, train the system for a relevant computer vision task, and use the latent features extracted from the trained network as visual representations. Although these visual encoders are highly optimised for the underlying vision tasks such as large-scale image classification or object detection~\citep{russakovsky-ilsvrc-2015}, it has been shown that the learned representations transfer very well into vision-to-language tasks such as image captioning~\citep{vinyals-show-2015, xu-show-2015}. Therefore,
the majority of IGT approaches rely on features extracted from state-of-the-art CNNs~\citep{simonyan-vgg-2014,ioffe-batch-2015,he-resnet-2016} trained for the ImageNet~\citep{deng-imagenet-2009} image classification task, where the output of the network is a distribution over 1000 object categories. These features usually come in two flavors~(Figure~\ref{fig:ict:cnnfeats}):
(i) spatial features which are feature maps $\spatfeats \in \R^{W\times H\times C}$ extracted from specific convolutional layers, and (ii) a pooled feature vector $\globfeats \in \R^{C}$ which is the outcome of applying a projection or pooling layer on top of spatial features. The main difference between these features is that the former is dense and preserves spatial information, while the latter is a compact, spatially-unaware representation. An even more compact representation is to use the posterior class probabilities ($\globfeats \in \R^{K}$) extracted from the output layer of a pretrained CNN, with $K$ denoting the size of the task-specific label set (for ImageNet, $K$ is 1000).
Finally, it is also possible to obtain \emph{a set of pooled feature vectors} (or \textit{local} features) from salient regions of a given image, with regions predicted by object detection CNNs \citep{girshick-rcnn-2014}.

\subsubsection{Sequence-to-sequence grounding with pooled features}
\label{sec:tasks:ict:cond}
\begin{figure}[t]
\centering
\includegraphics[width=.6\textwidth]{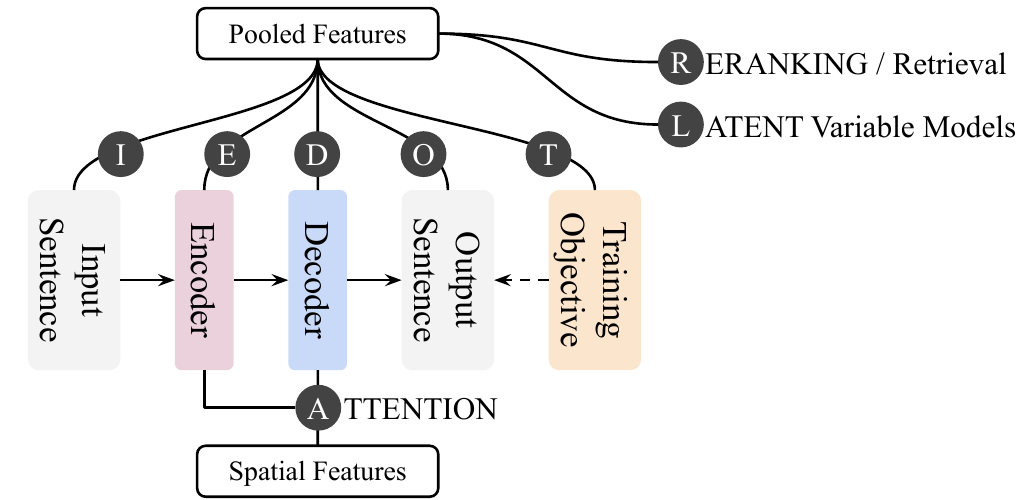}
\caption{A broad visualisation of the state of the art in image-guided translation.}
\label{fig:ict:approaches}
\end{figure}

The simplest and the most intuitive way of visually conditioning a sequence-to-sequence model is to employ pooled features in a way that they will interact with various components of the architecture.
These approaches are mostly inspired by the early works in neural image captioning~\citep{kiros-multimodal-2014,mao-deep-2014,vinyals-show-2015}, and are categorised in Figure~\ref{fig:ict:approaches} with respect to their entry points.

The very first attempt for neural image-guided translation comes from \cite{elliott2015multi}, where they formulate the problem as a semantic transfer from a source language model to a target language model, within an encoder-decoder framework without attention. They propose to initialise the hidden state(s) of the source language model~(LM), the target LM, or both, using pretrained VGG features~\citep{simonyan-vgg-2014}. Later initialisation variants are applied to attentive NMTs: \cite{calixto-elliott-frank:2016:WMT} and \cite{libovicky-cuni-2016} experiment with recurrent decoder initialisation while \cite{ma-EtAl:2017:WMT1} initialise both the encoder and the decoder, with features from a state-of-the-art ResNet~\citep{he-resnet-2016}. \cite{madhyastha-wang-specia:2017:WMT} explore the expressiveness of the posterior probability vector as a visual representation, rather than the pooled features from the penultimate layer of a CNN.

\cite{huang-attention-based-2016} take a different approach and enrich the source sentence representation with visual information by projecting the feature vector into the source language embedding space and then adding it to the beginning or the end of the embedding sequence. This allows the attention mechanism in the decoder to attend to a mixed-modality source representation instead of a purely textual one. Instead of the conventional ImageNet-extracted features, they make use of \textit{local features} from RCNN~\citep{girshick-rcnn-2014} to represent explicit visual semantics related to salient objects. In another model referred to as \textit{Parallel-RCNN}, they build five different source embedding sequences, each being enriched with a visual feature vector extracted from a different salient region of the image. A shared LSTM encodes these five sequences and average pools them to end up with the final source representation.

\cite{calixto-incorporating-2017} revisit the idea of source enrichment to extend it by simultaneously appending and prepending the projected visual features to the embedding sequence; and combining it with encoder and/or decoder initialisation. \cite{caglayan-lium-cvc-2017} explore different source and target interaction methods such as the element-wise multiplication between the visual features and the source/target word embeddings. \cite{delbrouck-dupont:2018:WMT} add another recurrent layer within the decoder in their \textit{DeepGRU} model, conditioned on the visual features and the bottom layer hidden state. Both recurrent layers simultaneously decide on the output probability distribution by additively fusioning their respective unnormalised logits.

As for transformer-based architectures, \cite{gronroos-memad-2018} revisit the source enrichment by adding the visual feature vector to the beginning of the embedding sequence~\citep{huang-attention-based-2016}.
They also experiment with modulating the output probability distribution through a time-dependent visual decoder gate. More interestingly, they explore different pooled visual representations such as scene--type associations~\citep{xiao-sun-2010}, action--type associations~\citep{yao-stanford-2011}, and object features from Mask R-CNN~\citep{he-maskr-2017}. 

\paragraph{\textbf{Multi-task learning.}}
Training an end-to-end neural model to perform multiple tasks at once can improve the model's task-specific performance by forcing it to exploit commonalities across the tasks involved~\citep{caruana1997multitask,dong-multitask-2015,luong2015multi}.
The \textit{Imagination} architecture, initially proposed by
\cite{elliott-imagination-2017} and later integrated into transformer-based NMTs by \cite{helcl-libovick-varis:2018:WMT},
attempts to leverage the benefits of multi-tasking by proposing a one-to-many framework which shares the sentence encoder between the translation task and an auxiliary visual reconstruction task. Besides the usual cross-entropy translation objective, the model weights are also optimised through a margin-based loss which minimises the distance between the ground-truth visual feature vector and the one predicted from the sentence encoding. The visual features are only used at training time and are not needed when generating translations.
\cite{zhou-visual-2018} further extends the \textit{Imagination} network by incorporating an attention\footnote{It should be noted that the attention here is over the source language encodings, and hence not a visual/spatial attention.} over source sentence encodings, with the query vector being the visual features. In this approach, the auxiliary margin-based loss is modified so that the output of the attention layer is considered a reconstruction of the pooled feature vector.


\paragraph{\textbf{Other approaches.}}
All grounding approaches covered so far rely on the maximum-likelihood estimation (MLE) principle for the sequence transduction task, \ie{}they try to maximise the log-probability of target sentences given the source sentences. \cite{zheng-EtAl:2018:WMT} extends MLE with a fine-tuning step, where they use reinforcement learning to find the model parameters which \textit{directly} maximise the translation metric BLEU. In terms of multimodality, they simply initialise the decoder with pooled features.
\citet{toyama-2016-latent}, \citet{calixto-2018-latent} and \citet{delbrouck-adversarial-2019} cast the problem as a latent variable model and resort to techniques such as variational inference and generative adversarial networks (GANs).
Finally, \citet{nakayama-zmt-2017} approach the problem from a \textit{zero-resource} perspective: they encode \{source caption, image\} pairs into a multimodal vectorial space using a max-margin loss. In a second step, they train the decoder using \{target caption, image\} pairs. Specifically, they do a forward-pass with the image as input and obtain the multimodal embedding, from which the recurrent decoder is trained to generate the target caption as usual. The image encoder is a pretrained VGG CNN. The \textit{zero-resource} aspect comes from the fact that the sets of pairs do not overlap \ie{}the approach does not require parallel IGT corpus.

\subsubsection{Visual attention}
\label{sec:tasks:ict:att}
Inspired by the previous success of visual attention in image captioning~\citep{xu-show-2015}, attentive approaches explore how to efficiently integrate a visual attention (approach A in Figure~\ref{fig:ict:approaches}) over the spatial features, alongside the language attention in NMTs. The most interesting research questions about visual attention are as follows: where to apply the visual attention, what kind of parameter sharing should be preferred and, how to fuse the output of language and visual attention layers.
\cite{caglayan-does-2016} and \cite{calixto-elliott-frank:2016:WMT} are the first works to tackle these questions, through a visual attention which uses the hidden state of the decoder as \textit{query} into the set of $W\times H$ spatial features. Their implementation is quite similar to the language attention, which results in two modality-specific contexts that should be fused before the output layer of the network. One notable difference is that \cite{caglayan-does-2016} experiment with a single multimodal attention layer \textit{shared} across modalities while \cite{calixto-elliott-frank:2016:WMT} keep the attention layers separate. Later on, \cite{caglayan-multimodal-2016} evaluate both \textit{shared} and \textit{separate} attentions with \textit{additive} and \textit{concatenative} fusion, and discover that proper feature normalisation is crucial for their recurrent approaches~\citep{caglayan-EtAl:2018:WMT}.
\cite{delbrouck-bilinearpooling-2017} propose a different fusion operation based on \textit{compact bilinear pooling}~\citep{fukui-mcb-2016}, to efficiently realise the computationally expensive outer product. Unlike \textit{additive} and \textit{concatenative} fusions, outer product ensures that each dimension of the language context vector interacts with each dimension of the visual context vector and vice-versa.
Follow-up studies extend the \textit{decoder-based visual attention} approach in different ways: \citet{calixto-doubly-2017} reimplement the \textit{gating mechanism}~\citep{xu-show-2015} to rescale the magnitude of the visual information before the fusion,
while \citet{libovicky-att-2017} introduce the \textit{hierarchical attention} which replaces the \textit{concatenative} fusion with a new attention layer that dynamically weighs the modality-specific context vectors.
Finally, \citet{arslan-doubly-2018} and \citet{libovicky-tfatt-2018} introduce the same idea into the Transformer-based~\citep{vaswani-attention-2017} architectures. Besides revisiting the \textit{hierarchical} attention, \cite{libovicky-tfatt-2018} also introduce \textit{parallel} and \textit{serial} variants. The former is quite similar to \cite{arslan-doubly-2018} and simply performs \textit{additive} fusion while the latter first applies the language attention, which produces the \textit{query} vector for the subsequent visual attention.
\cite{ive-etal-2019-distilling} extend \cite{libovicky-tfatt-2018} to add a 2-stage decoding process where visual features are only used in the second stage, through a visual cross-modal attention. They also experiment with another model where the attention is applied over the embeddings of object labels detected from the images.


In contrast to the \textit{decoder-based visual attention}, encoder-based approaches are relatively less explored. To that end, \cite{delbrouck-modulating-2017} propose conditional batch normalisation, a technique to modulate the batch normalisation layer~\citep{ioffe-batch-2015} of ResNet. Specifically, they condition the mean and the variance of the batch normalisation layer on the source sentence representation for informed feature extraction. In the same work,
\cite{delbrouck-modulating-2017} also propose to apply an \textit{early visual attention} inside the encoder, to yield inherently multimodal source encodings, on top of which the usual language attention would be applied by the decoder.

\subsubsection{Reranking and Retrieval based approaches}
\label{sec:tasks:ict:pipeline}
The most typical pipeline for MT is to obtain an \textit{n-best} list of translation candidates from an arbitrary MT system and select the best candidate amongst them after \textit{reranking} with respect to an aggregated score. This score is often a combination of several models that are able to quantitatively assess translation-related qualities of a candidate sentence, such as the adequacy or the fluency, for example. Each model is assigned a coefficient and an optimisation step is executed to find the best set of coefficients that maximise the translation performance on an held-out test set~\citep{och-mert-2003}.
The challenge for the IGT task is notably how to incorporate the visual modality into this pipeline in order to assign a better rank to visually plausible translations. To this end, \citet{caglayan-does-2016} combine a feed-forward language model~\citep{bengio-nplm-2003,schwenk-cslm-2006} and a recurrent NMT to rerank the translation candidates obtained from an SMT system. The language model is special in the sense that it is not only conditioned on \textit{n-gram} contexts but also on the pooled visual feature vector. In contrast, \citet{shah-wang-specia:2016:WMT} conjecture that the posterior class probabilities may be more expressive than a pooled representation for reranking, and treat each probability $\globfeats_{i}$ as an independent score for which a coefficient is learned.
In a recent work, \citet{lala-EtAl:2018:WMT} demonstrate that for the Multi30k dataset,
better translations are available inside an \textit{n-best} list obtained from a text-only NMT model, which allow up to 10 points absolute improvement in METEOR score.
They propose the multimodal lexical translation (MLT) model where they rerank the n-best list with scores assigned by a multimodal word sense disambiguation system based on pooled features.

Another line of work considers the task as a joint retrieval and reranking problem. \citet{hitschler-2016-multimodal} construct a multimodal/cross-lingual retrieval pipeline to \textit{rerank} SMT translation candidates. Specifically, they leverage a large corpus of target \{caption, image\} pairs, and retrieve a set of pairs similar to the translation candidates and the associated image. The visual similarity is computed using the Euclidean distance in the pooled CNN feature space. The initial translation candidates are then reranked with respect to their  -- inverse document frequency based -- relevance to the retrieved captions.
\citet{zhang-EtAl:2017:WMT2} also employ a combined framework of retrieval and reranking.
For a given \{caption, image\} pair, they first retrieve a set of similar training images. The target captions associated with these images are considered as candidate translations. They learn a multimodal word alignment between source and candidate words
and select the most probable target word for each source word. An \textit{n-best} list 
from their SMT is reranked using a bi-directional NMT trained on the aforementioned source/target word sequences. 
Finally, \citet{duselis-EtAl:2017:WMT} and \citet{gwinnup-EtAl:2018:WMT2} propose a pure retrieval system without any reranking involved. For a given image, they first obtain a set of candidate captions from a pretrained image captioning system. Two distinct neural encoders are  used to encode the source and the candidate captions, respectively. A mapping is then learned from the hidden space of the source encoder to the target one, allowing the retrieval of the candidate caption which minimises the distance with respect to the source caption representation.

\subsubsection{Comparison of approaches}
\label{sec:tasks:ict:comp}
Table~\ref{tbl:igt:sotacomp} presents BLEU and METEOR scores on the \lp{English}{German} \texttt{test2016} set of Multi30k dataset, as this is the test set that most studies report against. When possible, we annotate each score with the associated gain or loss with respect to the underlying unimodal MT baseline reported in the respective papers. The results concentrate around \textit{constrained} systems, which only allow the use of parallel Multi30k corpus during training. A few studies experiment with using external resources~\citep{calixto-doubly-2017,helcl-libovicky:2017:WMT,elliott-imagination-2017,gronroos-memad-2018} for pretraining the MT system and then fine-tuning it on Multi30k, or directly training the system on the combination of Multi30k and the external resource. Two such \textit{unconstrained} systems are also reported.

\begin{table}[t]
\centering
\caption{Automatic scores of state-of-the-art IGT methods on Multi30k \lp{English}{German} \texttt{test2016}: the table is clustered (and sorted by METEOR) across years for constrained systems, followed by unconstrained ones. Systems marked with ($\dagger$) are re-evaluated with tokenised sentences, \ff\,denotes the use of visual features other than ImageNet CNNs. The gains and losses are with respect to the MT baselines reported in the papers. The types refer to Figure~\ref{fig:ict:approaches}.}
\label{tbl:igt:sotacomp}
\renewcommand{\arraystretch}{1.1}
\resizebox{.99\textwidth}{!}{%
\begin{tabular}{@{}lccllc@{}}
\toprule
& \textbf{BLEU \uua} & \textbf{METEOR \uua} & \textbf{Type} & \textbf{Description} & \textbf{Arch.} \\ \midrule
  \citet{elliott2015multi}\dg               & 9.7 \na           & 24.7 \na          & E,D & Conditional LMs    & RNN   \\

  \citet{caglayan-does-2016}\dg             & 29.3 \ssmn{4.6}   & 48.5 \ssmn{4.3}   & A   & Shared Attention   & RNN   \\
  
  \citet{calixto-elliott-frank:2016:WMT}\dg & 28.8 \na          & 49.6 \na          & A   & Separate Attention & RNN   \\
  
  \citet{huang-attention-based-2016}\dg     & 36.8 \sspn{2.0}   & 54.4 \sspn{2.3}   &  I\ff  & Parallel RCNN-LSTMs & RNN \\
  
  \citet{hitschler-2016-multimodal}\dg      & 34.3 \na          & 56.0 \na          &  R  & Retrieval + Reranking & SMT \\

  \citet{toyama-2016-latent}                & 36.5 \sspn{1.6}   & 56.0 \sspn{0.7}   & L   & Variational            & RNN \\
  
  \citet{shah-wang-specia:2016:WMT}\dg      & 34.8 \sspn{0.2}   & 56.7 \sspn{0.1}   & R   & Visual Reranking      & SMT     \\
  \citet{caglayan-does-2016}\dg             & 36.2 \noc         & 57.5 \sspn{0.1}   & R   & Visual Reranking      & SMT     \\

  \midrule

  \citet{helcl-libovicky:2017:WMT}          & 31.9 \ssmn{2.7}   & 49.4 \ssmn{2.3}   & A   & Hierarchical Attention & RNN      \\

  \citet{calixto-incorporating-2017}        & 36.9 \sspn{3.2}   & 54.3 \sspn{2.0}   & I   & Input Prepend \& Append & RNN     \\
  
  \citet{calixto-doubly-2017}               & 36.5 \sspn{2.8}   & 55.0 \sspn{2.7}   & A   & Gated Attention    &  RNN    \\

  \citet{calixto-incorporating-2017}        & 37.3 \sspn{3.6}   & 55.1 \sspn{2.8}   & D   & Decoder Init.       &  RNN   \\

  \citet{elliott-imagination-2017}          & 36.8 \sspn{1.3}   & 55.8 \sspn{1.8}   & T   & Imagination        & RNN     \\
  
  \citet{caglayan-lium-cvc-2017}            & 38.2 \sspn{0.1}   & 57.6 \sspn{0.3}   & E,D  & Encoder Decoder Init.  & RNN    \\
                                            & 37.8 \ssmn{0.3}   & 57.7 \sspn{0.4}   & O    & Multiplicative Interaction & RNN \\
  
  \citet{delbrouck-modulating-2017}         & 40.5 \na          & 57.9 \na          & A    & Encoder Attention + CBN & RNN  \\
  \midrule

  \citet{arslan-doubly-2018}                & 41.0 \sspn{2.4}   & 53.5 \ssmn{1.5}  & A    & Parallel Attention  & Transformer   \\

  \citet{calixto-2018-latent}               & 37.6 \sspn{2.6}   & 56.0 \sspn{1.1}   & L    & Variational          & RNN \\
  \citet{helcl-libovick-varis:2018:WMT}     & 38.8 \sspn{0.7}   & 56.4 \sspn{0.2}   & T    & Imagination          & Transformer    \\
  \citet{libovicky-tfatt-2018}              & 38.5 \sspn{0.2}   & 56.5 \ssmn{0.2}   & A    & Hierarchical Attention & Transformer \\
                                            & 38.6 \sspn{0.3}   & 57.4 \sspn{0.7}   & A    & Parallel Attention     & Transformer \\
  \midrule
  \citet{ive-etal-2019-distilling}          & 38.0 \sspn{0.1}   & 55.6 \ssmn{0.3}   &D\ff  & 2-stage Decoder + Label Embs.        & Transformer \\
  \citet{libovicky-thesis-2019}             & 37.6 \sspn{0.9}   & 56.0 \sspn{0.9}   & A    & Hierarchical Attention  & RNN  \\
  \citet{caglayan-thesis-2019}              & 39.0 \sspn{0.1}   & 58.5 \sspn{0.1}   & E,D  & Encoder Decoder Init.   &  RNN \\
                                            & 39.4 \sspn{0.5}   & 58.7 \sspn{0.3}   & A    & Separate Attention + L$_2$ Norm.  & RNN \\
  \midrule
  \textbf{Unconstrained ensembles}      \\ \midrule
  \citet{helcl-libovick-varis:2018:WMT}     & 42.6 \sspn{2.2}   & 59.4 \sspn{0.4}   & T    & Imagination & Transformer            \\
  \citet{gronroos-memad-2018}               & 45.5 \noc         & \na               & I\ff & Input  Prepend & Transformer         \\

\bottomrule
\end{tabular}}
\end{table}

At a first glance, the automatic results reveal that (i) initially, neural systems were not able to surpass the SMT systems, (ii) the use of external resources is beneficial to boost the underlying baseline performance, which further manifests itself as a boost in the multimodal scores and (iii) careful tuning allows RNN-based models to reach and even surpass Transformer-based models.
From a multimodal perspective, the results are not very conclusive as there does not seem to be a single architecture, feature type or integration type that brings consistent improvements.
\cite{elliott2018adversarial} attempted to answer the question of how efficiently state-of-the-art models were integrating information from the visual modality and concluded that when models were adversarially challenged with wrong images at test time, the quality of the produced translations was not that much affected as one would expect. 
Later on, \cite{caglayan-probing-2019} showed how these seemingly insensitive architectures start to significantly rely on the visual modality, once words were systematically removed from source sentences during training and test. We believe that this latter finding may also be connected to the fact that better baselines benefit less from the visual modality~(Table~\ref{tbl:igt:sotacomp}) \ie sub-optimal architectures may leverage more from the visual information when compared to well trained NMT models. In fact, even the choice of vocabulary size may simulate \textit{systematic word removal}, if a significant portion of the source vocabulary are mapped to unknown tokens.
The same experimental pipeline of \cite{caglayan-probing-2019} also paved the way for assessing the particular strengths of some of the covered IGT approaches and showed that, the use of spatial features through visual attention is superior than initialising the encoders and the decoders using pooled features.

Lastly, if we take a look at the human evaluation rankings conducted throughout the WMT shared tasks, we see that the top three ranks for \lp{English}{German} and \lp{English}{French} are occupied by two unconstrained ensembles~\citep{gronroos-memad-2018,helcl-libovick-varis:2018:WMT}, the MLT Reranking~\citep{lala-EtAl:2018:WMT} and the DeepGRU~\citep{delbrouck-dupont:2018:WMT} systems in 2018. In 2017, the multiplicative interaction~\citep{caglayan-lium-cvc-2017}, unimodal NMT reranking~\citep{zhang-EtAl:2017:WMT2}, unconstrained \textit{Imagination}~\citep{elliott-imagination-2017}, encoder enrichment~\citep{calixto-incorporating-2017} and hierarchical attention~\citep{helcl-libovicky:2017:WMT} were ranked as top three, again for both language pairs.

    \subsection{Spoken language translation}
    \label{sec:slt}
    In spoken language translation, the non-text modality is the source language audio, which is translated into target language text. While source language transcripts may be available for training, at translation time the speech is typically the only input modality.
We begin this section with a brief introduction to speech-specific feature extraction (Section \ref{sec:slt:featureextraction}). Section~\ref{sec:slt:pipeline} reviews the current state of the art for the traditional pipeline methods and finally, Section~\ref{sec:slt:endtoend} covers the end-to-end methods which saw a rapid development in recent years.

\subsubsection{Feature extraction}
\label{sec:slt:featureextraction}
Even though many deep learning applications use raw input data, it is still common to use somewhat engineered features in speech applications.
The raw audio waveform consists of thousands of samples per second, and thus one-sample-at-a-time processing would be computationally very expensive. Instead, a \emph{spectrogram} representation is computed. It shows the signal activity at different frequencies, as a function of time. The frequency content is computed over frames of suitable length. The frame length trades off time and frequency precision: longer frames capture finer \emph{spectral} (\ie frequency) detail, but also describe a longer segment of time, which can be problematic as certain speech events (\eg the stop consonants \textipa{p}, \textipa{t}) can have a very short duration.

Next, a \emph{Mel-scale filterbank} 
is applied to each frame, and the logarithm of each filter's output 
is computed. This leads to \emph{log Mel-filterbank} features. The filterbank operation reduces the number of dimensions. However, these operations are also perceptually motivated: the filterbank by the masking of frequencies close to each other in the ear, the Mel-scale as it relates frequency to perceived pitch, and the logarithm by the relation of perceived loudness to signal activity \citep{pulkki2015communication}.

Continued efforts in learning deep representations from raw samples exist, with some success~\citep{sainath2015learning}. 
However, log Mel-filterbank vectors as input to deep neural network models~\citep{mohamed2012understanding} 
 remain the standard choice.
Additional, more complex features may be used to aid robustness to speaker variability~\citep{saon2013speaker} or recognition in tonal languages~\citep{ghahremani2014pitch}.

\subsubsection{State of the art in pipeline methods}
\label{sec:slt:pipeline}

Pipeline approaches in SLT chain together separate ASR and MT modules, and these naturally follow progress in their respective fields. A popular ASR system architecture is an HMM-DNN hybrid acoustic model~\citep{yu2017recent}, followed by an n-gram language model in the first decoding pass, and a neural language model for rescoring. This type of HMM-based ASR is essentially \emph{pipeline ASR}. In addition to pipeline ASR, \emph{end-to-end ASR} methods have recently gained popularity. Particularly, encoder-decoder architectures with attention have been successful, although on standard publicly available datasets HMM-based models still narrowly outperform end-to-end ones~\citep{luscher2019rwth}. \citet{chiu2018sota} show that encoder-decoder with attention ASR can outperform HMM-based models on an very large (12500h) proprietary dataset. Another common end-to-end ASR method is Connectionist Temporal Classification~(CTC) (\emph{e.g.}~\citet{li2019jasper}).

\citet{wang-sogou-tiic-2018} and \citet{liu2018ustc} place first and second, respectively, in the IWSLT 2018 evaluation campaign. Both apply similar pipeline architectures: a system combination of multiple different HMM-DNN acoustic models and LSTM rescoring for ASR, followed by a system combination of multiple Transformer NMT models for translation. \citet{liu2018ustc} additionally use an encoder-decoder with attention ASR to improve the system combination ASR results, although individually the end-to-end model is clearly outperformed by the HMM-DNN models. \citet{wang-sogou-tiic-2018} use an additional target-to-source NMT system for rescoring to improve adequacy. The systems also differ in interfacing strategies between ASR and MT.

In the latest IWSLT evaluation campaign in 2019, end-to-end SLT models were encouraged. However, the best performance was still achieved with a pipeline SLT approach, where \citet{pham2019iwslt} use end-to-end ASR and a Transformer NMT model. In the ASR module, an LSTM-based approach outperforms a Transformer model, though combining both in an ensemble proved beneficial. \citet{weiss-s2s-st-2017} and \citet{pino2019harnessing} also report competitive results using end-to-end ASR, with \citet{pino2019harnessing} surpassing the state-of-the-art in SLT. End-to-end ASR has attracted attention in SLT, because it allows for parameter transfer in end-to-end SLT (\emph{e.g.}~\citet{berard-e2e-audiobooks-2018}, and  Figure~\ref{fig:slt:approaches:e2e:a}).

\begin{table}[t]
    \centering
    \caption{SLT formulated as Bayesian search, for \mbox{translation $y$}, \mbox{source language transcript $z$}, \mbox{source language speech $x$}, and \mbox{set of all possible transcripts $\mathrm{Z}$}.}
    \label{tab:pipeline-slt-search}
    {\renewcommand{\arraystretch}{2}
    \begin{tabular}{ll@{}}
    \toprule
        \textbf{End-to-end search}          & $\argmax_{y}P(y|x)$ \\
        \textbf{General pipeline search}    & $\argmax_{y}\sum_{z\in\mathrm{Z^\prime}(x)}P(y|z)P(z|x)$ \\
        \textbf{Pure serial pipeline}       & $\mathrm{Z^\prime}(x) = \big\{\argmax_{z}P(z|x)\big\}$ \\
        \textbf{Loosely coupled pipeline}   & $\mathrm{Z^\prime}(x) \subset \mathrm{Z}$ \\
        \textbf{Tightly coupled pipeline}   & $\mathrm{Z^\prime}(x) = \mathrm{Z}$ \\
    \bottomrule
    \end{tabular}
    }
\end{table}

\paragraph{\textbf{Challenges in pipeline SLT}}

Research in pipeline SLT has specifically focused on the interface between ASR and MT. There is a clear mismatch between MT training data and ASR output, caused by the ASR noise characteristics (\ie transcription errors), and the ASR output dissimilarity with respect to the written text due to lack of capitalisation and punctuation, and the disfluencies (\eg repetitions and hesitations), which naturally occur in speech.
\citet{ruiz2014assessing,ruiz2015phonetically,ruiz2017assessing}
quantify the effect of ASR errors on MT. In a linear mixed-effects model, the amount of WER added on top of gold standard transcripts has a direct effect on TER increase. 
The results do not vary over different ASR systems. Minor localised ASR errors can result in longer distance errors or duplication of content words in NMT. Homophonic substitution error spans (\eg \textit{anatomy $\rightarrow$ and that to me}) are shown to account for a significant portion of ASR errors and to have a large impact on translation quality.
With regards to noise robustness, it is noted that the utterances which were best translated by phrase-based MT, had higher average WER than utterances which were best translated by NMT. In general, NMT has been established as particularly sensitive to noisy inputs~\citep{belinkov2018synthetic,cheng-etal-2018-towards}.

One approach to address the mismatch is training the MT system on noisy, ASR-like input. \citet{peitz2012spoken} use an additional phrase-table trained on ASR-outputs on the SLT corpus. \citet{tsvetkov-etal-2014-augmenting-translation} augment a phrase-table with plausible ASR misrecognitions. These errors are synthesised by mapping each phrase to phones via a pronunciation dictionary, and randomly applying heuristic phone-level edit operations. 

\citet{sperber2017toward} first train an NMT system on reference transcripts, and then fine-tune on noisy transcripts. The noise is sampled from a uniform distribution over insertions, deletions or substitutions, with optional unigram weighting for the substitutions and insertions. Additionally, a deletion-only noise is used. Smaller amounts of noise are shown to improve SLT results, but increasing noise levels to actual test-time ASR levels (rather high, at~40\%) only degrades performance. Increased noise is noted to produce shorter outputs, which in turn are punished by the BLEU brevity penalty. A precision-recall tradeoff is observed: the system could either drop uncertain inputs (better precision) or try to guess translations (better recall). Fine-tuning with deletion-only noise biases the system to produce longer outputs, which is shown to counteract the effect of noisy inputs producing shorter outputs. \citet{pham2019iwslt} use the data augmentation method SwitchOut~\citep{wang-etal-2018-switchout}, to make their NMT models more robust to ASR errors. During training, SwitchOut randomly replaces words in both the source and the target sentences.

Another approach to cope with the mismatch is to transform the ASR-output into written text.
\citet{wang-sogou-tiic-2018} apply a Transformer-based punctuation restoration and heuristic rules which remove disfluencies and transform written out numbers and quantities into numerals. \citet{liu2018ustc} experiment with NMT-based transformations in both directions: producing ASR-like text from written text for training the translation system, or producing written text from ASR-like text as a test-time bridge between ASR and translation. Transforming the MT training data into an ASR-like format consistently outperforms inverse normalization of ASR-output, though both are beneficial in the final system combination.

Long audio streams typically need to be segmented into manageable length pieces using voice activity detection~\citep{ramirez2007vad}, or more elaborate speaker diarisation methods~\citep{anguera2012diarization}.
These methods may not produce clean sentence boundaries.
This is a clear problem in MT, as the boundaries can cut between actual sentences. \citet{liu2018ustc} alleviate the problem by applying an LSTM-based resegmenter after the ASR system. \citet{pham2019iwslt} combine resegmentation, and casing and punctuation restoration into a single ASR post-processing task, and apply an NMT model.

\paragraph{\textbf{Coupling between ASR and MT}}

The SLT search is often described in Bayesian terms as shown in Table~\ref{tab:pipeline-slt-search}. Generally, pipeline search is based on the assumption that $P(y|z,x) = P(y|z)$, \ie given the source language transcript, the translation does not depend on the speech. It is still possible to take the uncertainty of the transcription into account under this conditional independence assumption, but it rules out the use of paralinguistic cues, \eg prosody.
In \emph{pure serial} pipeline search, first the 1-best ASR result is decoded, then only this 1-best result is translated. The hard choice in 1-best decoding is especially susceptible to error propagation. Early work in SLT found consistent improvements with \emph{loosely coupled} search, where a rich representation carrying the ASR uncertainty, such as an N-best list or word lattice, is used in translation. \emph{Tightly coupled} search, \ie \emph{joint decoding}, is also possible, although the application is limited by excessive computational demands. In tightly coupled search, the translation model would also influence which ASR hypotheses were searched further. This was done by representing both the ASR and the phrase-based MT search spaces as Weighted Finite State Transducers (WFST).~\citep{matusov2006integrating,zhou2013smt}

\citet{osamura2018using} implement a type of loose coupling by using the softmax posterior distribution from the ASR module as the input for NMT. Loose coupling via using lattices as input in NMT is not straightforward. \citet{sperber-etal-2017-neural} implement LatticeLSTM for lattice inputs in RNN-based NMT, and find that  preserving the uncertainty in the ASR output is beneficial for SLT. \citet{zhang-etal-2019-lattice} further propose a Transformer model which can use lattice inputs, and find that it outperforms both a standard Transformer and a LatticeLSTM baseline in an SLT task. However, tight coupling of NMT and ASR has not been proposed in pipeline SLT.

In addition to coupled decoding, end-to-end SLT leverages coupled training. This can avoid suboptimization; for phrase-based MT and HMM-GMM ASR, \citet{he2011wer} show how optimizing the ASR component purely for WER can produce worse results in SLT. \citet{he2013speech} foreshadow end-to-end neural SLT systems, proposing a joint, end-to-end optimization procedure for a pipeline of HMM-GMM ASR and phrase-based MT. In the proposed approach, the ASR and MT components are first trained separately, and then the whole pipeline is jointly optimized for sentence-level BLEU, by iteratively sampling sets of competing hypotheses from the pipeline and updating the parameters of the submodels discriminatively.

\subsubsection{End-to-end spoken language translation}
\label{sec:slt:endtoend}


The first attempts to use end-to-end methods for SLT were published in 2016.
This period saw experimentation with a wide variety of approaches,
before research focus converged on sequence-to-sequence architectures.
These early methods
\citep{duong-etal-2016-attentional,anastasopoulos-etal-2016-unsupervised,bansal-etal-2017-towards}
were able to align source language audio to target language text,
but they were not able to perform translation.
The first true end-to-end SLT system is presented by \citet{berard2016listen}.
Still a proof-of-concept,
it was trained on BTEC \lp{French}{English} with synthetic audio containing a small number of speakers.


%

\begin{figure}[t]
    \centering
    \includegraphics[width=0.55\textwidth]{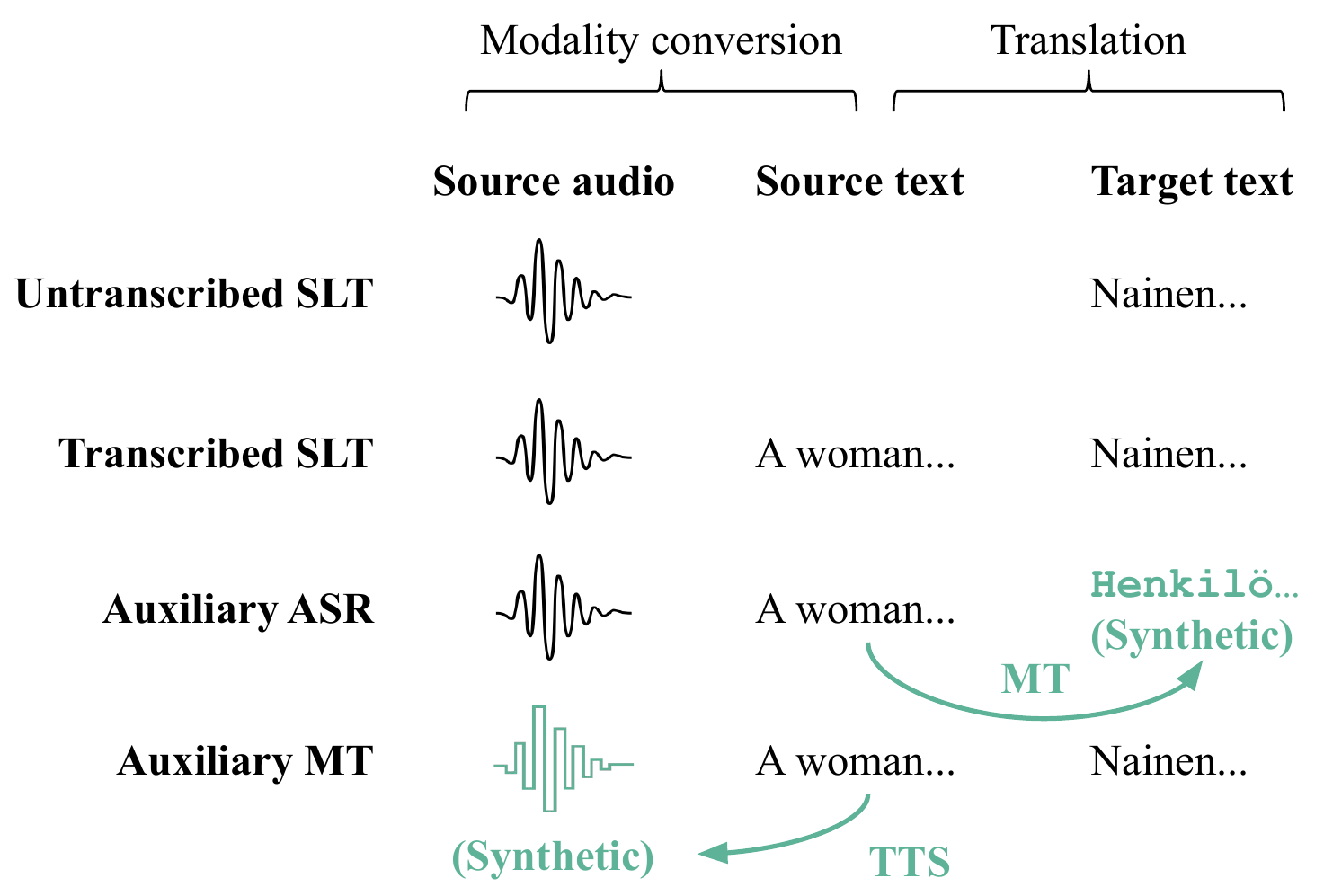}
    \caption{%
Four types of data that can be used to train SLT systems. Untranscribed SLT is the minimal type of data for end-to-end systems. Adding source text transcripts completes the triple. The source text is an intermediate representation which divides the SLT mapping into a modality conversion and a translation. Two types of auxiliary data, ASR and MT data, form adjacent pairs in the triple, leaving one of the ends empty. The auxiliary data can be used as is for pretraining or multi-task learning, or it can be completed into synthetic triples using external TTS or MT systems.
    \label{fig:slt:typesofdata}}
\end{figure}

\begin{figure}[t]
    \centering
    \includegraphics[width=\textwidth]{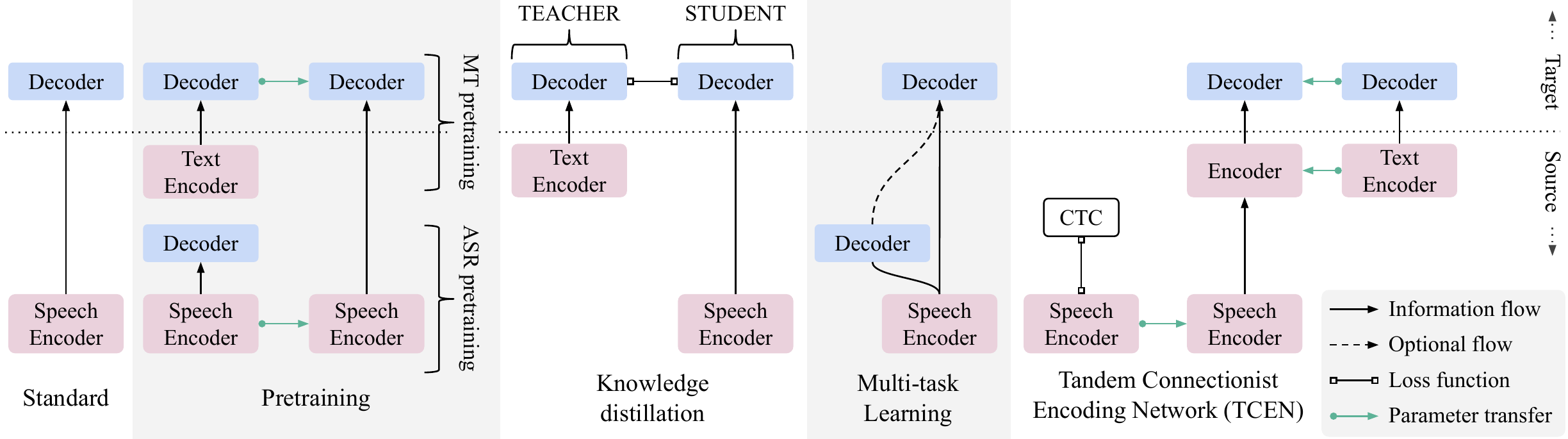}
    \caption{Learning setups for end-to-end SLT: The \textit{standard} framework uses untranscribed SLT data. Auxiliary data can be exploited in different ways such as by \textit{pretraining} the encoder through ASR, \textit{pretraining} the decoder through MT, \textit{knowledge distillation}, or \textit{multi-task learning}. The optional link in \textit{multi-task learning} results in 2-step decoding. TCEN combines multiple types of pretraining.
    \label{fig:slt:approaches:e2e:a}}
\end{figure}

Figure \ref{fig:slt:typesofdata} shows the different types of training data applicable for SLT.
The standard learning setup for end-to-end SLT is only able to train from untranscribed SLT data.
The task is very challenging, as data of this type is scarce, and the representation gap between source audio and target text is large.
The source transcript is useful as an intermediary representation, a stepping stone to divide the gap into two smaller ones:
modality conversion and translation.
Many learning setups (see Figure \ref{fig:slt:approaches:e2e:a}),
e.g. pretraining, multi-task learning, and knowledge distillation,
have been applied for exploiting the source transcripts.
In early experiments,
no new examples are introduced for the auxiliary task(s);
Only source transcript labels for the SLT examples were added.
Later the same learning setups have been applied to exploit more abundant auxiliary ASR and MT data.

An important milestone towards parity with pipeline approaches was to achieve better translation quality when both the end-to-end system and the pipeline system are trained on the same SLT data.
This milestone was reached by \citet{weiss-s2s-st-2017},
training on the 163h Fisher\&Callhome \lp{Spanish}{English} data set. As pipeline methods are naturally capable of exploiting the more abundant paired ASR and MT data, but in this case
this condition was unrealistically constrained.
When the constraint is lifted, pipeline methods improve to a level that is difficult or impossible to reach on small amounts of source audio-translated text data.
The effective use of auxiliary data 
was a key insight going forward towards achieving parity with pipeline approaches.

Figure \ref{fig:slt:approaches:e2e:a}
shows learning setups that have been applied for exploiting source transcripts and auxiliary data.
\citet{weiss-s2s-st-2017} use a multi-task learning procedure with ASR as the auxiliary task, training only on transcribed SLT data.
In multi-task learning~\citep{caruana1997multitask}, multiple tasks are trained in parallel, with some network components shared between the tasks.
%
%
\citet{berard-e2e-audiobooks-2018} compare \emph{pretraining} (sequential transfer) with \emph{multi-task learning} (parallel transfer),
finding very little difference between the two.
In pretraining, some of the parameters from a network trained to perform an auxiliary task are used to initialise parameters in the network for the main task.
The system is trained only on transcribed SLT data,
with two auxiliary tasks:
pretraining the encoder and decoder with ASR and textual MT respectively.
\citet{stoian2019analyzing} compare the effects of pretraining on auxiliary ASR datasets of different languages and sizes, concluding that the WER of the ASR system is more predictive of the final translation quality than language relatedness.

\citet{anastasopoulos-chiang-2018-tied} make the line between pipeline and end-to-end approaches more blurred by using a multi-task learning setup with two-step decoding.
First the source transcript is decoded using the ASR decoder.
A second SLT decoder attends to both the speech input and the hidden states of the ASR decoder.
While the system is trained end-to-end, the two-step decoding is still necessary at translation time.
The system is trained only on transcribed SLT data.
%
\citet{liu2019end} focus on exploiting source transcripts by means of knowledge distillation.
They train the student SLT model to match the output probabilities of a text-only MT teacher model,
finding that knowledge distillation is better than pretraining.
\citet{inaguma2019espnet} also see substantial improvements from knowledge distillation when adding auxiliary textual parallel data.
%
\citet{wang2019bridging} introduce the \emph{Tandem Connectionist Encoding Network}~(TCEN),
which allows neural network components to be pretrained while
minimising both the number of parameters not transferred from the pretraining phase,
and the mismatch of components between pretraining and finetuning.
The final network consists of four components: ASR encoder, MT encoder, MT attention and MT decoder.
The ASR encoder is pretrained with a Connectionist Temporal Classification objective function,
which does not require a separate ASR decoder which would go to waste after pretraining.
The last three parts can be pretrained with a textual MT task.

\citet{jia2019leveraging} show that \emph{augmenting auxiliary data} is more effective than multi-task learning.
MT data is augmented with synthesised speech, while ASR data is augmented with synthetic target text by forward translation using a text-only MT system (see Figure~\ref{fig:slt:typesofdata}).
These kinds of synthetic data augmentation are conceptually similar to the highly successful practice of
using backtranslation~\citep{sennrich2016improving} to exploit monolingual data in textual MT.
With both pretraining and multi-task learning, the end-to-end system slightly outperforms the pipeline.
Adding synthetic data substantially outperforms the pipeline.
The systems are both trained on exceptionally large proprietary corpora: 
ca 1300 h translated speech and 49000 h transcribed speech.
Controversially the system is also evaluated on a proprietary test set.
The speech encoder is divided into two parts, of which only the first is pretrained on an ASR auxiliary task.
The entire decoder is pretrained on the text MT task.
%
\citet{pino2019harnessing} evaluate several pretraining and data augmentation approaches.
They use TTS to synthesise source audio for parallel text data,
finding that the effect depends on the quality and quantity of the synthetic data.
Using textual MT to synthesise target text from ASR data is clearly beneficial.
Pretraining the speech encoder on an ASR task is useful for the lower resourced \lp{English}{Romanian}, but not for \lp{English}{French}.
Pretraining on ASR is not a good substitute for using textual MT for augmenting the ASR data,
but does speed up convergence of the SLT model.
Using a combination of a VGG Transformer speech encoder and decoder,
they very nearly reach parity with a strong pipeline system.


\citet{bansal2019pre} apply \emph{crosslingual pretraining}, by pretraining on high-resource ASR to improve low-resource SLT.
They use a small \lp{Mboshi}{French} SLT corpus without source transcripts.
As Mboshi has no official orthography, transcripts may be difficult to collect.
Pretraining the speech encoder using a completely unrelated high-resource language, English,
effectively allows to account for acoustic variability, such as speaker and channel differences.
%
\citet{gangi2019onetomany} train a one-to-many multilingual system to translate from English
to all 8 target languages of the \mbox{MuST-C} corpus,
with an additional task pair for English ASR.
Prepending a target language tag to the input~\citep{johnson-googles-2016}, is not effective in multilingual SLT,
resulting in many acceptable translations into the wrong language.
Better results are achieved with a stronger language signal using \textit{merge},
a language-dependent shifting operation.
%
\citet{inaguma2019multilingual} train multilingual models for
\{\textsc{en, es}\} $\rightarrow$ \{\textsc{en, fr, de}\} SLT.
They achieve better results with the multilingual models than with bilingual ones,
including pipeline methods for some test sets.

Noise-based data augmentation methods have also been applied to the speech audio.
\citet{bahar2019usingSpecAugment} and \citet{digangi2019data} apply spectral augmentation (SpecAugment),
which randomly masks blocks of features that are consecutive in time and/or frequency.

\begin{table}[t]
\centering
\caption{BLEU scores for SLT methods on \lp{English}{French} Augmented LibriSpeech/test.\\
All systems are end-to-end,
except for the pipeline system marked with a dagger (\dg).
\label{tbl:slt:librispeech}}
\renewcommand{\arraystretch}{1.1}
\begin{tabular}{@{}lccccl@{}}
\toprule
\textbf{Approach}                  & \textbf{BLEU \uua} & \multicolumn{3}{c}{\textbf{Training data}} & \textbf{Description} \\ 
\cmidrule(lr){3-5}
                                   &                    & SLT (h) & ASR (h) & MT (sent) & \\
\midrule
\citet{berard-e2e-audiobooks-2018} & 13.4               & 100h    &         &           & CNN+LSTM. Multi-task. \\
\citet{di2019adapting}             & 13.8               & 236h    &         &           & CNN+Transformer. \\
\citet{bahar2019usingSpecAugment}  & 17.0               & 100h    & 130h    & 95k       & Pyramidal LSTM. Pretraining, augmentation. \\
\citet{liu2019end}                 & 17.0               & 100h    &         &           & Transformer. Knowledge distillation. \\
\citet{inaguma2019multilingual}    & 17.3               & 472h    &         &           & CNN+LSTM. Multilingual. \\
\citet{pino2019harnessing}         & 21.7               & 100h    & 902h    & 29M       & CNN+Transformer. Pretraining, augmentation. \\
\citet{pino2019harnessing}\dg      & 21.8               & 100h    & 902h    & 29M       & End-to-end ASR. CNN+LSTM. \\
\bottomrule
\end{tabular}
\end{table}

\subsubsection{End-to-end SLT architectures}

There is a large variety of architectures that have been applied to end-to-end SLT,
with no clear favourite having emerged.
However, recent architectures all follow some type of sequence-to-sequence architectures that makes use of attention mechanisms.

Two varieties of LSTM layers have been used:
standard bi-LSTM \citep[\emph{e.g.}][]{jia2019leveraging}
and pyramidal bi-LSTM \citep[\emph{e.g.}][]{duong-etal-2016-attentional,berard2016listen,bahar2019usingSpecAugment}.
The pyramidal construction of the encoder downsamples the long speech input sequence,
making subsequent bi-LSTM layers and the attention mechanism faster and alignment easier.
\citet{berard2016listen} use convolutional attention, finding it to be particularly useful with long input sequences.
Following \citet{weiss-s2s-st-2017}, \citet{berard-e2e-audiobooks-2018} move away from the pyramidal bi-LSTM encoder architecture to convolution followed by bi-LSTM.
The prepended convolutional layers perform the downsampling of the audio signal, making the pyramidal construction unnecessary.

Transformers have also been used in many SLT systems.
\citet{liu2019end} propose an architecture in which all encoders and decoders are standard Transformer encoders and decoders respectively.
\citet{pino2019harnessing} further prepend VGG-style convolutional blocks to Transformer encoders and decoders, in order to replace the positional embedding layer of the standard Transformer architecture and to downsample the signal.
\citet{gangi2019onetomany} use a speech encoder which begins with stacks of convolutional layers interleaved with 2D self-attention~\citep{dong2018speech},
followed by a stack of Transformer layers.
\citet{salesky2019fluent} revisit the network-in-network~\citep{lin2013network} architecture to achieve downsampling: parameters are shared spatially in a similar way to CNN, but a full multi-layer perceptron network is applied to each window. 

Convolutional Neural Networks are used in many SLT architectures,
but only in combination with LSTM or Transformer, not in isolation.
The combined CNN-LSTM architecture is popular in end-to-end ASR \citep{watanabe2018espnet}.
The CNN is well suited for reduction of the time scale to something manageable,
and modeling short range dependencies.
The appended LSTM or Transformer
is useful for encoding the semantic information for translation.
%
The CNNs used in SLT are typically 2D convolutions (parameter sharing across both time and frequency).
Time Delay Neural Networks (TDNN) are still popular in ASR, 
but have not to the best of our knowledge been used in end-to-end SLT. 
TDNNs can be seen as a 1D convolution, only sharing parameters across time.
The VGG~\citep{simonyan-vgg-2014} architecture of CNNs is used in SLT, but not ResNet~\citep{he-resnet-2016}.

\paragraph{\textbf{Comparison of architectures.}}
In SLT, the choice between LSTM and Transformer architectures doesn't seem to be a settled matter:
recent papers use both.
Both architectures are powerful enough, when stacked into sufficiently deep networks.
\citet{pino2019harnessing} present a result in favour of the Transformer,
as they only reach parity with their pipeline using Transformers, but not LSTMs.
\citet{inaguma2019espnet} find that Transformers consistently outperform LSTMs in their experiments.
A downside of LSTM is slow training on the very long sequences encountered in speech translation.
While the Transformer parallelises to a larger extent, making training fast,
it is not immune to long sequences, as the self-attention is quadratic in memory w.r.t. the length.
The Transformer also lacks explicit modelling of short range dependencies,
due to the self-attention learning dependencies of any range with equal difficulty.
\citet{di2019adapting} attempt to augment the Transformer to alleviate some of its shortcomings.

\paragraph{\textbf{Decoding units.}}
In textual NMT, subword-level decoders have become the standard choice~\citep{sennrich2016neural}.
Most end-to-end SLT systems use character-level decoders.
Although word level decoding is rare,
\citet{bansal2018low} focus on a low-computation setting,
deciding to use word-level decoding to shorten the sequence length.
Some well-performing recent systems use subword units \citep{liu2019end,jia2019leveraging,pino2019harnessing,bansal2019pre}.
\citet{wang2019bridging} find characters to work better than subwords in their system.

\paragraph{\textbf{Has parity with pipeline approaches been reached?}}
Recent results \citep{jia2019leveraging,pino2019harnessing} show that on certain tasks
with large enough datasets of high-quality,
end-to-end systems can reach the same or even better performance than pipeline systems.
In low-resource settings, end-to-end systems do not perform as well.
However, in the IWSLT 2019 evaluation campaign~\citep{niehues-iwslt-2019},
the pipeline system of \citet{schneider2019kit} clearly outperforms all end-to-end submissions.
%
\citet{sperber2019attention} find that current methods do not use auxiliary data effectively enough.
The amount of transcribed SLT data is critical: When the size of the data containing all three
of source audio, source text and target text is sufficient, end-to-end methods outperform pipeline methods.
In lower resource settings where the amount of SLT data is insufficient, pipeline methods are better.

Table \ref{tbl:slt:librispeech} shows results on the \lp{English}{French} Augmented LibriSpeech test set, which is one of the most competed test sets for SLT, particularly end-to-end SLT. It shows the rapid increase in performance during the last two years, and the importance of maximally exploiting available training data.



\section{Future Directions}
\label{sec:future}
The previous sections provide a detailed overview of resources, definitions of various kinds of multimodal MT, and the extensive work that has been devoted to develop models for the different tasks. However, multimodal MT 
is still in its infancy. This is especially the case for truly end-to-end models, which have only appeared in recent years. Future work should explore more realistic settings that go beyond restricted domains and rather artificial problems such as visually-guided image caption translation.

\subsection{Datasets and resources}

Image-guided translation has, thus far, been studied with small-scale datasets \citep{elliott-multi30k:-2016}, and there is a need for larger-scale datasets that bring the resources for this task closer to the size of image captioning \citep{chen-microsoft-2015} and machine translation datasets \citep{tiedemann-parallel-2016}. Larger-scale datasets have started to appear for video-guided translation~\citep{sanabria-how2:-2018,wang-vatex-2019}. Spoken-language translation datasets \citep{kocabiyikoglu-librispeech-slt-2018, niehues-iwslt-2018} are smaller than standard automatic speech recognition datasets. A common challenge in multimodal translation is the need for crosslingually aligned resources, which are expensive to collect \citep{elliott-multi30k:-2016}, or can result in a small dataset of \textit{clean} examples \citep{kocabiyikoglu-librispeech-slt-2018}. Future work will obviously benefit from larger datasets, however, researchers should further explore the role of data augmentation strategies \citep{jia2019leveraging} in both spoken language translation and visually-guided translation.

\subsection{Evaluation and ``verification''}

A significant challenge in image-guided translation has been to demonstrate that a model definitively improves translation with image guidance. This has resulted in more focused evaluation datasets that test noun sense disambiguation~\citep{elliott-findings-2017,lala-multimodal-2018} and verb sense disambiguation~\citep{gella-etal-2019-cross}. In addition to new evaluations, researchers are focusing their efforts on determining whether image-guided translation models are sensitive to perturbations in the inputs. \citet{elliott2018adversarial} showed that the translations of some trained models are not affected when guided by incongruent images (\ie{the translation models were not guided by the image that the source language sentence describes, instead they are guided by a randomly selected image; see Section~\ref{sec:tasks:ict:comp} for more details}); \citet{caglayan-probing-2019} demonstrated that training models with masked tokens increases the sensitivity of models to incongruent image guidance; and, more recently, \citet{dutta-chowdhury-elliott-2019-understanding} showed that trained models are more sensitive to textual perturbations than incongruent image guidance. Overall, there is a need for more focused evaluations, especially in a wider variety of language pairs, and for models to be explicitly evaluated in these more challenging conditions. Future research on visually-guided translation should also ensure that new models are actually using the visual guidance in the translation process.

In spoken language translation, this line of research into focused evaluations might involve digging into the cases where a good transcript is not enough to disambiguate the translation. One possible case is translating into a language where the speaker's gender matters, such as French or Arabic~\citep{elaraby2018gender}.
End-to-end SLT systems have the potential to use non-linguistic information from the speech signal to tackle these challenges, but it is currently unknown to which extent they are able to do so. 

\subsection{Shared tasks}

In addition to stimulating research interest, shared task evaluation campaigns enable easier comparison of results by encouraging the use of standardised data conditions.
The choice of data condition can be made with many aims in mind.
To set up a race for state-of-the-art results using any and all available resources,
it is enough to define a common test set.
For this goal, any additional restrictions are unnecessary or even detrimental.
For example the GLUE natural language understanding task~\citep{wang-etal-2018-glue} takes this approach.

On the other hand, if the goal is to achieve as fair as possible comparison between architectures, then strict limitations on the training data are required as well.
Most evaluation campaigns choose this approach. However, it is far from trivial to select an appropriate set of data types to include in the condition. In many tasks, the use of auxiliary or synthetic data has proved vitally useful,
e.g. exploiting monolingual data in textual MT using backtranslation~\citep{sennrich2016improving}. In spoken language translation, the use of auxiliary data has prompted some discussion
of when end-to-end systems are considered to have reached parity with pipeline systems.
To answer this question in a fair comparison, both types of systems should be evaluated under standardised data conditions.

\subsection{\textit{Multi}modality and new tasks}

Most previous work on multimodal translation emphasises multimodal \textit{inputs} and unimodal outputs, mainly text. The integration of speech synthesis, and also a better integration of visual signals in generated communication is required for improved intelligent systems and interactive artificial agents. In addition to multimodal outputs, there should be a stronger emphasis on real-time language processing and translation. This new emphasis would also result in a closer integration of models for spoken language translation models and visually-guided translation.

In SLT, the visual modality could contribute both complementary and disambiguating information. In addition, visual speech recognition, automatic lip reading in particular~\citep[\eg{}][]{chung2017lip}, could aid SLT for example in audio noise robustness. The 
How2 dataset should allow a flurry of research in the nascent field of audio-visual SLT. \citet{wu2019transformer} present exploratory first results. BLEU improvements over the best non-visual baseline are not found, although the visual modality improves results when comparing between model using cascaded deliberation.

In zero-shot translation, a multilingual model is used for translating between a language pair that was not included in the parallel training data~\citep{firat-etal-2016-zero,johnson-googles-2016}.
For example, if a model does zero-shot \lp{French}{Chinese} translation,
the training data contains language pairs with French as the source language 
and Chinese as the target language 
but no parallel \lp{French}{Chinese} data.
Considering ongoing research into multilingual translation models also in multimodal translation~\citep[\eg{}][]{inaguma2019multilingual},
and the fact that multimodal translation training data of sufficient size is available for a very limited number of language pairs,
we expect an interest in zero-shot multimodal language translation in the future.

\section{Conclusions}
\label{sec:conclusions}

Multimodal machine translation provides an exciting framework for further development in grounded cross-lingual natural language understanding combining work in NLP, computer vision and speech processing. This paper provides a thorough survey of the current state of the art in the field focusing on specific tasks and benchmarks that drive the research. This survey details the essential language, vision, and speech resources that are available to researchers, and discusses the models and learning approaches in the extensive literature on various multimodal translation paradigms. Combining these different paradigms into truly multimodal end-to-end models of natural cross-lingual communication will be the goal of future developments, given the foundations laid out in this survey.


\section*{Acknowledgments}
\label{sec:acknowledgments}
%
%
%

This study has been supported by the MeMAD project, funded by the European Union's Horizon 2020 research and innovation programme~(grant agreement~\textnumero{}~780069), the FoTran and MultiMT projects, funded by the European Research Council~(ERC) under the European Union's Horizon 2020 research and innovation programme~(grant agreements~\textnumero{}~771113 and~\textnumero{}~678017 respectively), and the MMVC project, funded by the Newton Fund Institutional Links grant programme~(grant ID 352343575). We would also like to thank Maarit Koponen for her valuable feedback and her help in establishing our discussions of machine translation evaluation.


\bibliographystyle{spbasic}
\bibliography{surveynourl}

\end{document}